\pdfoutput=1

\documentclass[11pt]{article}

\usepackage{EMNLP2022}

\usepackage{times}
\usepackage{latexsym}
\usepackage{bm}
\usepackage{graphicx}
\usepackage{multirow}
\usepackage{booktabs}
\usepackage{subfig}
\usepackage{amsmath}

\usepackage[T1]{fontenc}

\usepackage[utf8]{inputenc}

\usepackage{microtype}

\usepackage{enumitem}
\usepackage[compact]{titlesec}
\usepackage{inconsolata}

\expandafter\def\expandafter\normalsize\expandafter{%
    \normalsize
    \setlength\abovedisplayskip{3pt}
    \setlength\belowdisplayskip{5pt}
    \setlength\abovedisplayshortskip{3pt}
    \setlength\belowdisplayshortskip{5pt}
}

\newtheorem{innercustomthm}{Corrolary}
\newenvironment{customthm}[1]
  {\renewcommand\theinnercustomthm{#1}\innercustomthm}
  {\endinnercustomthm}

\title{A Multilingual Perspective Towards the Evaluation of Attribution Methods in Natural Language Inference}

\author{Kerem Zaman\thanks{~~Work done while at Boğaziçi University.} \\
  UNC Chapel Hill \\
  \texttt{kzaman@cs.unc.edu} \\\And
  Yonatan Belinkov\thanks{~~Supported by the Viterbi Fellowship in the Center for Computer Engineering at the Technion.} \\
  Technion – Israel Institute of Technology \\
  \texttt{belinkov@technion.ac.il} \\}

\begin{document}
\maketitle
\begin{abstract}
Most evaluations of attribution methods focus on the English language. In this work, 
we present a multilingual approach for evaluating attribution methods for the Natural Language Inference (NLI) task in terms of   faithfulness and plausibility. 
First, we introduce a novel cross-lingual strategy to measure faithfulness based on word alignments, which eliminates the drawbacks 
of erasure-based evaluations.
We then perform a comprehensive evaluation of attribution methods, considering different output mechanisms and aggregation methods.
Finally, we augment the XNLI dataset with highlight-based explanations,  providing a multilingual NLI dataset with highlights, to support future exNLP studies. Our results show that attribution methods performing best for plausibility and faithfulness are different.\footnote{Our code and data are available at \url{https://www.keremzaman.com/explaiNLI}.}

\end{abstract}

\section{Introduction}

The opaqueness of large pre-trained models  like BERT \citep{devlin-etal-2019-bert} and GPT \citep{Radford2018ImprovingLU} motivates developing  explanation methods \citep{Wallace2020InterpretingPO}, which aim to attribute importance to particular input features \citep{Springenberg2015StrivingFS,LRP,Ribeiro2016WhySI,integrated-grads}, such as words in a textual input. Two main criteria for evaluating such methods are plausibility and faithfulness \citep{jacovi-goldberg-2020-towards}. Plausibility can be defined as the consistency between explanations and human expectations, while faithfulness is defined as the consistency between explanations and the model's underlying decision-making process.

Prior evaluations of attributions along these dimensions \citep{atanasova-etal-2020-diagnostic, deyoung-etal-2020-eraser, ding-koehn-2021-evaluating} suffer from several limitations. First, they have been limited in the range of considered attribution methods and the mechanism of calculating the attributions. Second, 
standard faithfulness evaluations, such as erasure-based ones \citep{deyoung-etal-2020-eraser}, entail running the model on  examples outside of the training distribution \citep{bastings-filippova-2020-elephant}. Third, prior plausibility evaluations  are limited to English-only datasets due to the lack of multilingual datasets with highlighted explanations.

In this work, we aim to fill these gaps.  Our main contribution is a new framework for evaluating the faithfulness of attribution methods.  Inspired by \citet{jacovi-goldberg-2020-towards}'s criterion for faithful explanations as giving similar explanations for similar inputs, we propose to use cross-lingual sentences (translations) as similar inputs. Given a multilingual model, we argue that faithful attributions should point to words that are aligned in two translations of the same sentence. This approach avoids out-of-distribution inputs by utilizing cross-lingual sentences as \emph{naturally ocurring} input perturbations. 

We focus on Natural Language Inference (NLI) as a case study, since it is a central task that has been widely used as a test bed for attribution methods \citep{atanasova-etal-2020-diagnostic, deyoung-etal-2020-eraser, Jain2019AttentionIN, kim-etal-2020-interpretation, wiegreffe-marasovic-2021-review, prasad-etal-2021-extent}. 
We compare eight attribution methods, including different mechanisms of computation varying the output and the aggregation of input feature importance scores.

First, 
we experiment with the cross-lingual XNLI dataset \citep{conneau2018xnli}, multilingual BERT \citep{devlin-etal-2019-bert}, and XLM-R \citep{Conneau2020UnsupervisedCR}, and discover large differences in the faithfulness of different attribution methods. 
Second, we find that certain attributions are more plausible and that the choice of computation mechanism has a large effect in some cases. As far as we know, this is the first comprehensive study investigating the effect of different types of outputs when evaluating attributions.

Informed by our comprehensive evaluation, we augment the multilingual XNLI dataset \citep{conneau2018xnli} with highlight-based explanations by extracting highlights for the English part of  XNLI and projecting along word alignments to other languages. We perform a plausibility evaluation with the resulting dataset, which we dub e-XNLI, and perform a human evaluation on a subset of the dataset to validate its adequacy.

Finally, when comparing the ranking of attribution methods by plausibility and faithfulness, we find that no single method performs best. Different methods have different pros and cons, and may therefore be useful in different scenarios.  In summary, this work provides: 
\begin{itemize}[itemsep=2pt,parsep=2pt,topsep=2pt]
\item A novel faithfulness evaluation framework. 
\item A comprehensive evaluation of attribution methods, which may guide practitioners when applying such methods.
\item A dataset containing explanations in multiple languages for the NLI task, which may support future multilingual exNLP studies.
\end{itemize}

\section{Background}
\subsection{Properties for Evaluating Attributions}

Many properties have been defined to evaluate explanations with respect to different aspects, such as plausibility and faithfulness \citep{jacovi-goldberg-2020-towards}, sufficiency \citep{deyoung-etal-2020-eraser}, stability and consistency \citep{RobnikSikonja2018PerturbationBasedEO}, and confidence indication \citep{atanasova-etal-2020-diagnostic}. As two prominent ones, we focus on faithfulness and plausibility. 

\subsubsection{Faithfulness} \label{sec:background-faithfulness}

Faithfulness is the measure of how much an interpretation overlaps with the reasoning process of the model. In other words, if the scores given by an attribution method are compatible with the decision process behind the model, the interpretation is considered faithful. Such compatability may be instantiated in different ways. For instance, \citet{ding-koehn-2021-evaluating} measure faithfulness through model consistency and input consistency. For model consistency, they compare attribution scores of a given model and its distilled version. For input consistency, they compare attribution scores of perturbed input pairs. 

Perturbing inputs or erasing parts of the input is a widely-used technique for faithfulness evaluation \citep{Arras2017WhatIR, serrano-smith-2019-attention, deyoung-etal-2020-eraser, ding-koehn-2021-evaluating, atanasova-etal-2020-diagnostic}. The basic idea is to observe the effect of changing or removing parts of inputs on model output. For instance, if removing words with high attribution scores changes the model output, then the explanation is faithful. For these methods, the change in prediction score is usually assumed to be caused by deletion of the significant parts from the input. However, the main reason might be the out-of-distribution (OOD) inputs created by the perturbations \citep{bastings-filippova-2020-elephant}. The dependence on  perturbations that result in OOD inputs is the main drawback of common faithfulness evaluation methods.
In Section \ref{sec:faithful} we propose a new evaluation that overcomes this drawback.

\subsubsection{Plausibility}

Plausibility is a measure of how much an explanation overlaps with human reasoning \citep{ding-koehn-2021-evaluating}. In particular, if an attribution method gives higher scores to the part of the inputs that affect the decision according to humans, then it is plausible. Typically, human-annotated highlights (parts of the input) are used for plausibility evaluation \citep{wiegreffe-marasovic-2021-review}, which we also follow in this work. However, some recent studies use lexical agreement \citep{ding-koehn-2021-evaluating}, human fixation patterns based on eye-tracking measurements \citep{hollenstein-beinborn-2021-relative}, and machine translation quality estimation \citep{Fomicheva2021TranslationED}.

\subsection{Overview of Attribution Methods}

In this work, we focus on the evaluation of local post-hoc methods, which provide explanations to the output of a model for a particular input by applying additional operations to the model's prediction \citep{danilevsky-etal-2020-survey}. Local post-hoc methods can be grouped into three categories: methods based on gradients, perturbations, or simplification \citep{atanasova-etal-2020-diagnostic}. In gradient-based methods, the gradient of the model's output with respect to the input is used in various ways for calculating attribution scores on the input. Perturbation-based methods   calculate  attribution scores according to the change in the model's output after perturbing the input in different ways. Simplication-based methods simplify the model to assign attributions. For instance, LIME \citep{Ribeiro2016WhySI} trains a simpler surrogate model covering the local neighborhood of the given input. Other post-hoc methods outside of these 3 categories  \citep{kokhlikyan2020captum} include Layer Activation \citep{Karpathy2015VisualizingAU}, which uses activations of each neuron in the output of a specific layer, and NoiseTunnel \citep{Smilkov2017SmoothGradRN, Adebayo2018LocalEM}.

\looseness=-1
The attribution methods we evaluate are: InputXGradient \citep{Shrikumar2016NotJA}, Saliency \citep{Simonyan2014DeepIC}, GuidedBackprop \citep{Springenberg2015StrivingFS}, and IntegratedGradients \citep{integrated-grads} as gradient-based methods;  Occlusion \citep{Zeiler2014VisualizingAU} and Shapley Value Sampling \citep{Ribeiro2016WhySI} as perturbation-based;
LIME \citep{Ribeiro2016WhySI} as simplification-based; and 
 Layer Activation \citep{Karpathy2015VisualizingAU}.
Details about these methods appear in Appendix \ref{sec:appendix-atrributions}. 

\subsection{Output Mechanisms and Aggregation Methods}

Most previous studies compute attributions when the output is the top prediction score.  More formally, let $f(\textbf{x}^{(i)})$ denote the output of a classification layer, where $x^{(i)}$ is $i$-th instance of the dataset. Then, the score of the top predicted class can be expressed $\max_{} f(\textbf{x}^{(i)})$. We also compare with the case when the output is the loss value calculated with respect to the gold label. For the common cross-entropy loss, the loss output can be expressed as $y^{(i)} log(f(\textbf{x}^{(i)}))$ where $y^{(i)}$ is the gold label. Furthermore, some attribution methods, such as InputXGradient and Saliency, return importance scores for each dimension of each input word embedding, which need to be aggregated to obtain a single score for each word. While prior studies use different aggregation operations, namely mean and $L_2$, we examine their effect exhaustively.

Denote the importance score for the $k$-th dimension of the $j$-th word embedding of $\textbf{x}^{(i)}$  as $u^{(i)}_{jk}$. Then we obtain an attribution score per word, $\omega^{(i)}_{\textbf{x}_j}$, using mean aggregation as follows:
\begin{equation}
\omega^{(i)}_{\textbf{x}_j} = \frac{1}{d} \sum_{k=0}^d u^{(i)}_{jk}
\end{equation}
where $d$ is the number of dimensions for the embedding.
Similarly, we define the attribution score per word using $L_2$ aggregation as follows:
\begin{equation}
\omega^{(i)}_{\textbf{x}_j} = \sqrt{ \sum_{k=0}^d (u^{(i)}_{jk})^2 }.
\end{equation}

\subsection{Natural Language Inference}

Natural Language Inference (NLI) is a well-established Natural Language Understanding (NLU) task where the objective is deciding the relation between given sentence pairs \citep{Consortium96usingthe, condoravdi-etal-2003-entailment, bos-markert-2005-recognising, Dagan2005ThePR, maccartney-manning-2009-extended, poliak-2020-survey}. When a sentence pair is given, namely a premise and a hypothesis, there are 3 possible outcomes: (i) premise entails hypothesis; (ii) premise and hypothesis contradict; or (iii) they are neutral. This setting makes the task suitable to be modeled as a text classification task. 

Although there are many human-annotated NLI datasets, we focus on the MNLI \citep{Williams2018ABC}, XNLI \citep{conneau2018xnli} and e-SNLI \citep{Camburu2018eSNLINL} datasets. MNLI is a collection of 433K sentence pairs from 10 genres of written and spoken English where pairs are labeled as \emph{entailment}, \emph{contradiction} or \emph{neutral}. This dataset is also part of a general NLU benchmark called GLUE \citep{Wang2018GLUEAM}. XNLI is the cross-lingual extension of the MNLI dataset in which sentence pairs from the validation and test sets of MNLI are translated into 15 languages. the e-SNLI dataset is the enhanced version of SNLI  \citep{Bowman2015ALA}, an English-only NLI dataset having the same format as MNLI, with human-annotated explanations in the form of highlights.

\begin{figure*}[t]
    \includegraphics[width=\textwidth]{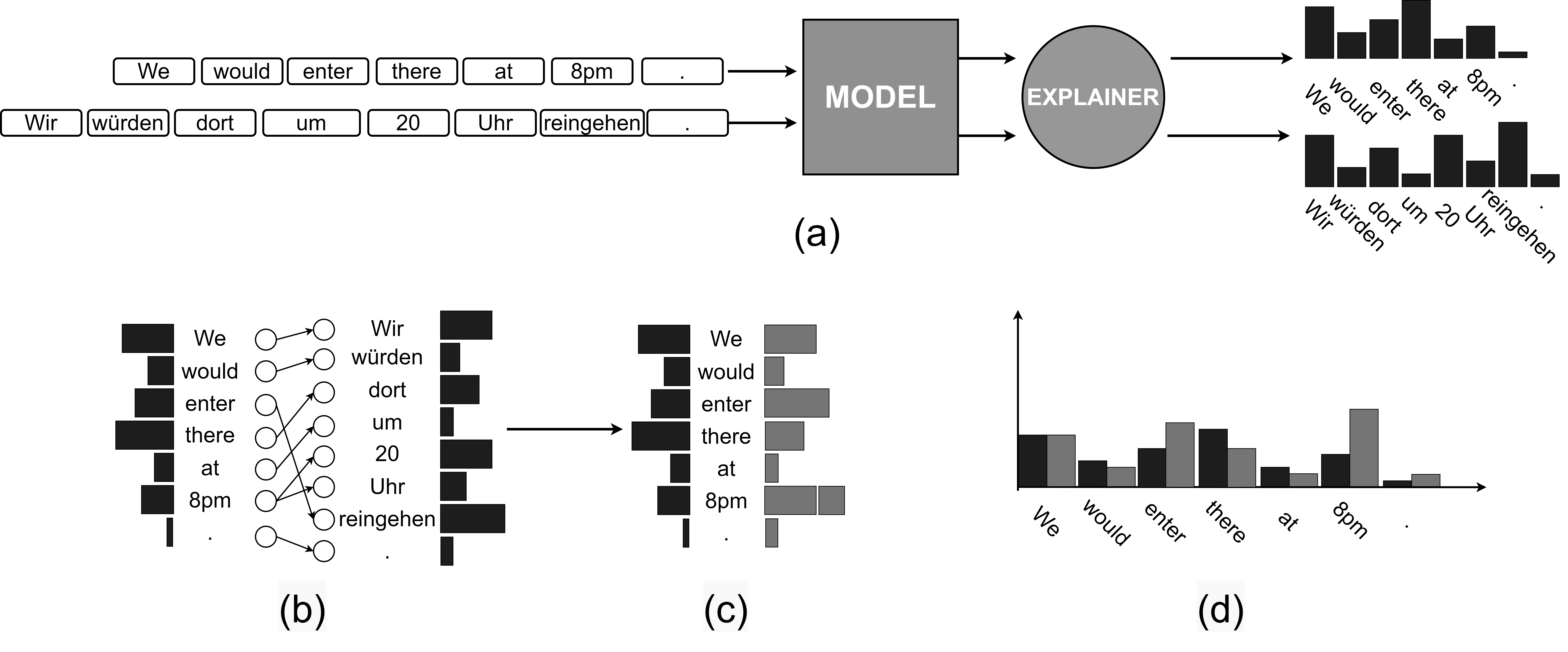}
    \caption{Illustration of cross-lingual faithfulness evaluation. (a) For any en--XX sentence pair (in this example, English--German), we pass each item of the pair through the cross-lingual model and attribution method, to get attribution scores. (b) We extract word alignments by using \texttt{awesome-align} and (c) align scores for the words in German with the ones in the English language by summing the scores of corresponding German words for each English word. (d) Finally, we get two different distributions for the English sentence: the calculated attribution scores and the aligned attribution scores. We compare them to evaluate faithfulness.}
    \label{crosslingual-faithfulness}
\end{figure*}

\section{Faithfulness} \label{sec:faithful}
\subsection{Evaluation Methods}

\subsubsection{Crosslingual Faithfulness Evaluation} \label{sec:multi-faithful}

In faithfulness evaluation, erasure-based methods examine the drop in prediction scores by removing the important tokens from the input (Section~\ref{sec:background-faithfulness}). However, the drop in the prediction scores may be the result of the altered, out-of-distribution inputs \citep{bastings-filippova-2020-elephant}.   To overcome this problem, we design a new strategy to evaluate faithfulness by relying on cross-lingual models and datasets. Before diving into details, let us recall Corrolary \ref{twopointone} from  \citet{jacovi-goldberg-2020-towards}.
\begin{customthm}{2}\label{twopointone}
  An interpretation system is unfaithful if it provides different interpretations for similar inputs and outputs.
\end{customthm}

The key intuition behind our method is to use translation pairs to provide similar inputs to a single model. In particular, we assume a multilingual model that can accept inputs from different languages, such as multilingual BERT (mBERT; \citealt{devlin-etal-2019-bert}). Then, we can examine the attribution scores of matching parts (words or phrases) of the similar inputs.\footnote{We investigate the similarity of translation pairs via their multilingual representations in Appendix \ref{sec:input-similarity}, finding that translation pairs do form similar inputs for a multilingual model.}

This idea consists of several steps. First,  construct multiway translation pairs of which source and targets are English and another languages, respectively. Second,  calculate attribution scores for instances in English and other languages. Third, align the attribution scores  between source and target through word alignments. Finally, correlate attribution scores computed for English instances  with the ones for corresponding words in other languages.  By looking at the correlation between corresponding parts of the inputs, we measure how consistent the model is for similar inputs. 
Figure \ref{crosslingual-faithfulness} illustrates the cross-lingual faithfulness evaluation procedure.

More formally, let $\mathbf{x}^{(i)}_{c} = \langle x^{(i)}_{c, 1}, x^{(i)}_{c, 2}, \ldots , x^{(i)}_{c, n} \rangle$ denote the $i$-th instance of the dataset for language $c$ (out of $C$ languages), where $x^{(i)}_{c, j}$ stands for the \mbox{$j$-th} word of the instance. Let $A = \{ (x^{(i)}_{{en}, k}, x^{(i)}_{c, j}) :  x^{(i)}_{{en}, k} \in \mathbf{x}^{(i)}_{en}, x^{(i)}_{c, j} \in \mathbf{x}^{(i)}_{c}  \}$ be the set of words from $\mathbf{x}^{(i)}_{c}$ that are aligned with words in the corresponding English  sentence, $\mathbf{x}^{(i)}_{en}$.\footnote{We use English as the reference language since our cross-lingual model performs best on it and since the word aligner we use was originally fine-tuned and evaluated on en--XX language pairs.} Denote by  $\omega^{(i)}_{{x}_{c, j}}$ the attribution score for word $x^{(i)}_{c, j}$ and let $\omega^{(i)}_{\mathbf{x}_c} = \langle \omega^{(i)}_{{x}_{c, 1}}, \omega^{(i)}_{{x}_{c, 2}}, \ldots, \omega^{(i)}_{{x}_{c, n}}  \rangle$. In order to align attribution scores for instances from another language with the English ones, we define the aligned attribution score for each word in the reference language as the sum of the attribution scores of the corresponding words in the target language:
\begin{equation}\overline{\omega}^{(i)}_{{x}_{c, k}} = \sum_{(x^{(i)}_{{en}, k}, x^{(i)}_{{c}, j} ) \in A}^{} \omega^{(i)}_{{x}_{c, j}} \end{equation}  
 By aligning scores, we obtain equivalent attribution scores in the target language for each word in the source language.
 For the example in Figure~\ref{crosslingual-faithfulness}, we have $\overline{\omega}^{(i)}_{\text{8pm}} = \omega^{(i)}_{\text{20}} + \omega^{(i)}_{\text{Uhr}}$, because  $\{(\text{8pm}, \text{20} ), (\text{8pm}, \text{Uhr} )\} \subset A$.

Finally, we define the cross-lingual faithfulness ($\rho$) of a dataset as the average Spearman correlation between attribution scores for English and aligned attribution scores for all other languages:
\begin{equation} \label{eq:rho}
 \rho = \frac{1}{C-1} \frac{1}{M} \sum_{c \neq {en}}^{} \sum_{i=0}^M \rho_{\omega^{(i)}_{\mathbf{x}_{en}}, \overline{\omega}^{(i)}_{\mathbf{x}_{c}} }
\end{equation}

\looseness=-1
The main advantage of this approach is in avoiding the OOD problem: Translation pairs form naturally occurring perturbations that are part of the model's training distribution, unlike the synthetic inputs formed by erasure-based methods. We also reduce  language-specific bias by using translations of the same sentence in different languages. Furthermore, our approach provides a grayscale notion of faithfulness, as advocated by  \citet{jacovi-goldberg-2020-towards}.

\subsubsection{Erasure-based Faithfulness Evaluation}

We compare our method with erasure-based faithfulness evaluation metrics, namely sufficiency and comprehensiveness \citep{deyoung-etal-2020-eraser}.  
We stick to \citeauthor{deyoung-etal-2020-eraser}'s definitions and choices along the experiments.

Let $m(\textbf{x}^{(i)})_j$ be the model output of the $j$-th class for the $i$-th data point and $r^{(i)}$ be the most important tokens to be erased, decided according to attribution scores. Comprehensiveness measures  the drop in  prediction probability after removing the important tokens  (higher values are better):

\begin{equation}
\!\!\textrm{comprehensiveness} \!=\! m(\textbf{x}^{(i)})_j \!-\! m(\textbf{x}^{(i)} \backslash r^{(i)})_j
\end{equation}

Sufficiency measures the drop when only the important tokens are kept (lower values are better):

\begin{equation}
\textrm{sufficiency} = m(\textbf{x}^{(i)})_j - m(r^{(i)})_j
\end{equation}

$r^{(i)}$ is the top-$k_d$ words according to their attribution scores, where $k_d$ depends on the dataset. However, choosing an appropriate $k$ can be tricky, especially when human rationales are not available to decide an average length. Also, the variable $k_d$ makes scores incomparable across datasets. To solve these issues, \citeauthor{deyoung-etal-2020-eraser} propose Area Over Perturbation Curve (AOPC) metrics for sufficiency and comprehensiveness, based on bins of tokens to be deleted. They calculate comprehensiveness and sufficiency when deleting the top tokens in each bin, and obtain AOPC metrics by averaging the scores for each bin.
Here we group the top 1\%, 5\%, 10\%, 20\%, 50\% tokens into bins in the order of decreasing attribution scores. 

\subsection{Faithfulness Experiments}

\paragraph{Experimental setup}
We use the XNLI dataset \citep{conneau2018xnli} to construct translation pairs where source and target are English and other languages, respectively. We use \texttt{awesome-align} \cite{dou-neubig-2021-word} to align attribution scores for the corresponding words in translation pairs.\footnote{We use the model provided by the authors, which was multilingually fine-tuned without consistency optimization, due to its good zero-shot performance. We examine the effect of word alignments in Appendix \ref{sec:crosslingual-faithfulness-ablation}.} We fine-tune mBERT and XLM-R\textsubscript{base} for English on the MNLI dataset \citep{N18-1101} with 3 different seeds for each. For cross-lingual faithfulness evaluation, we only use the languages that are common in the top-5 languages for both types of cross-lingual models(when performing zero-shot prediction on non-English languages). This gives Bulgarian, German, Spanish and French (C = 5). The cross-lingual performance of our models on all XNLI languages appears in Appendix \ref{sec:appendix-mbert}.

\subsubsection{Cross-lingual Faithfulness Experiments}
\label{sec:crosslingual-faithfulness-experiments}

\begin{table}[t]
\centering
\begin{tabular}{lrr}
\toprule
\multirow{2}[3]{*}{\textbf{Method}} & \multicolumn{2}{c}{\textbf{$\rho$}} \\
\cmidrule(lr){2-3}
 & \textbf{TP} & \textbf{Loss} \\
\midrule
InputXGradient ($\mu$) & .0588 & .0756 \\
InputXGradient ($L_2$) & \textbf{.7202} & \textbf{.7208} \\
Saliency ($\mu$) & .5676 & .5680 \\
Saliency ($L_2$) & .5664 & .5670 \\
GuidedBackprop ($\mu$) & .0026 & .0020 \\
GuidedBackprop ($L_2$) & .5664 & .5670 \\
IntegratedGrads ($\mu$) & .1878 & .2439 \\
IntegratedGrads ($L_2$) & .6095 & .5636 \\
Activation ($\mu$) & .5552 & .5552 \\
Activation ($L_2$) & .6965 & .6965 \\
LIME & .0421 & .0677 \\
Occlusion & .1480 & .2049 \\
Shapley & .2283 & .2742 \\
\bottomrule
\end{tabular}
\caption{
Cross-lingual faithfulness results: Average correlations measured for different attribution methods on the XNLI dataset. Scores are averaged across all models including different architectures and seeds. Attributions are performed with respect to the top prediction (TP) score and the loss. InputXGradient with $L_2$ aggregation is the best performing method in both cases.
}
\label{crosslingualfaithfulness-results}
\end{table}

\begin{figure}[t!]
    \includegraphics[width=\columnwidth]{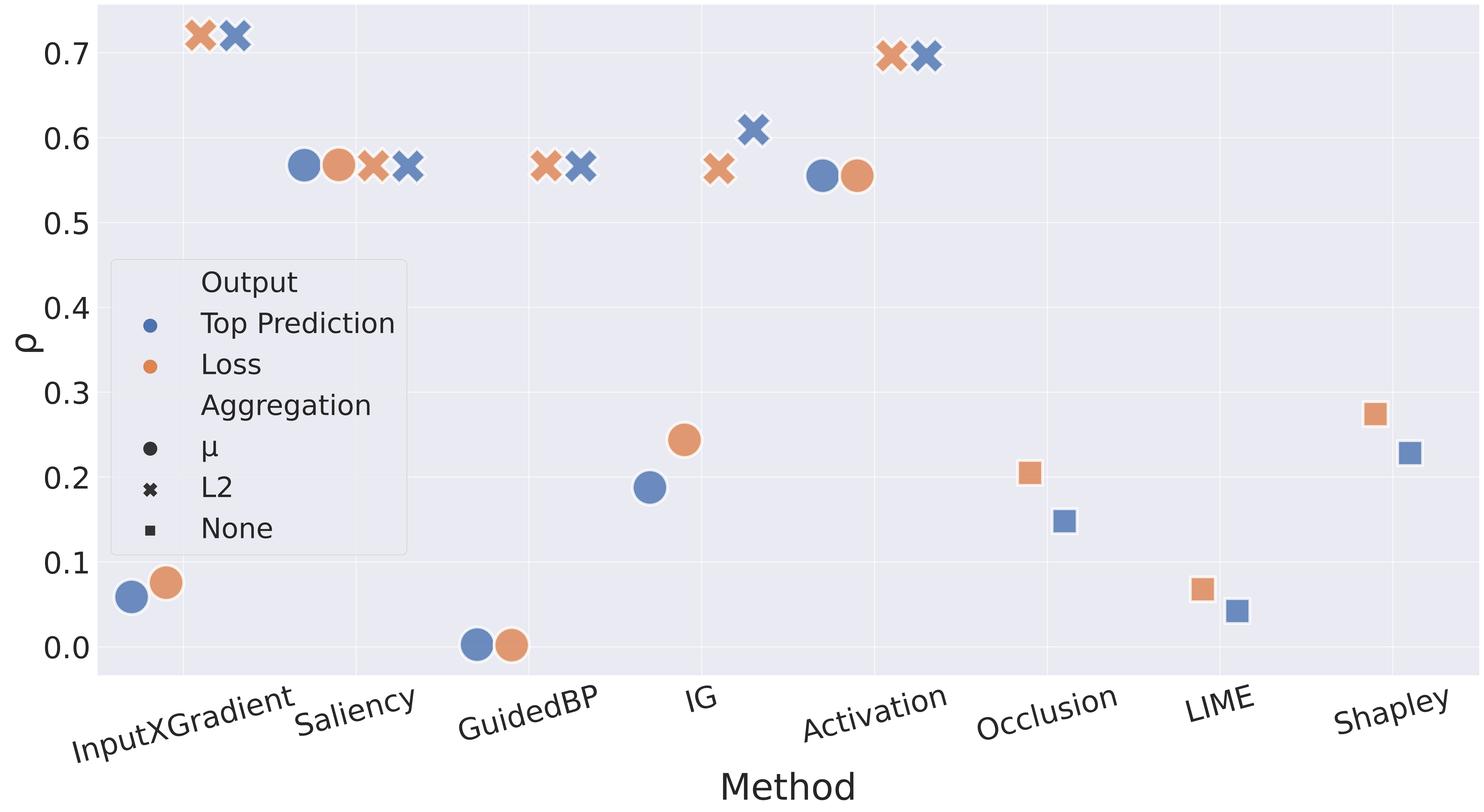}
    \caption{Comparison of cross-lingual faithfulness along output and aggregation dimensions.  $L_2$ mostly outperforms mean ($\mu$) aggregation and calculations with respect to the loss are the same as or slightly better than ones with respect to the top prediction score.}
    \label{cf_results_fig}
\end{figure}

Table \ref{crosslingualfaithfulness-results} shows   cross-lingual faithfulness results for each attribution method,  when computing attributions with regard to top prediction or loss, and when aggregating input scores with $L_2$ or mean aggregation.   The results exhibit a large variation, indicating that our cross-lingual faithfulness evaluation is able to expose differences between attribution methods. 
InputXGradient with $L_2$ aggregation is the most faithful attribution method for both types of attribution calculation.
We also observe that gradient-based attribution methods (first 8 rows in Table~\ref{crosslingualfaithfulness-results}) usually generate more faithful explanations than perturbation-based ones (last two rows), in line with prior work \citep{atanasova-etal-2020-diagnostic}.

Figure \ref{cf_results_fig} shows the effect of  aggregation methods and output mechanisms on cross-lingual faithfulness. In all cases, $L_2$ aggregation outperforms  mean aggregation by large margins, except for Saliency, where mean aggregation is slightly better than $L_2$ aggregation. Since Saliency returns the absolute value, which is analogous to $L_1$ aggregation, the exception in the results makes sense. Considering output mechanisms, attribution scores calculated with respect to loss are more faithful than ones calculated with respect to the top prediction score in almost all cases. For Integrated Gradients with $L_2$ aggregation and GuidedBackprop with mean aggregation, calculating attribution scores with respect to the top prediction score performs better.

Recall that our cross-lingual faithfulness measure averages correlations across languages (Eq.~\ref{eq:rho}). 
To analyze the effect of languages, especially the ones that are poorly represented by multilingual models, we repeat the same experiments with the worst-performing 3 languages: Thai, Swahili, and Urdu.  Table \ref{crosslingualfaithfulness-results-per-lang} shows correlations per language when averaged across all combinations of methods, outputs and aggregations. The results show little variation across top-performing languages. When the relation between NLI performance and faithfulness is considered, it turns out there is a strong correlation between them (Pearson correlation coefficient and p-value are as follows: $r=0.83$,  $p = 0.02$) and poorly represented languages yield lower faithfulness scores.
Detailed results per language and attribution method are given in Appendix~\ref{sec:crosslingual-faithfulness-results-per-lang}.

\begin{table}[h]
\centering
\begin{tabular}{l c c c c | c c c }
\toprule
  & bg & de & es & fr & th & sw & ur \\ 
  \midrule 
 $\rho$ & $.36$  &   $.38$ & $.41$ & $.40$ & .14 & .27 & .25 \\ 
 \textbf{Acc} & $.73$ & $.74$ & $.77$ & $.76$ & $.63$ & $.58$ & $.62$  \\ 
\bottomrule
\end{tabular}
\caption{
Cross-lingual faithfulness results ($\rho$) per language averaged across all attribution methods on the XNLI dataset, and NLI accuracies for comparison. 
}
\label{crosslingualfaithfulness-results-per-lang}
\end{table}

\subsubsection{Erasure-based Faithfulness Experiments}

\begin{table}[t]
\small
\centering
\begin{tabular}{lrr rr}
\toprule
\multirow{2}[3]{*}{\textbf{Method}} & \multicolumn{2}{c}{\textbf{comp.} $\uparrow$} & \multicolumn{2}{c}{\textbf{suff.} $\downarrow$} \\
\cmidrule(lr){2-3} \cmidrule(lr){4-5}
 & \textbf{TP} & \textbf{Loss} & \textbf{TP} & \textbf{Loss} \\
\midrule
InputXGradient ($\mu$) & .2945  & .3072  & .2812  & .2784 \\
InputXGradient ($L_2$) & .3146  & .2980  & \textbf{.2479}  & .2682 \\
Saliency ($\mu$) & .3075  & .3017  & .2588  & .2584 \\
Saliency ($L_2$) & \textbf{.3158}  & .3010  & .2640  & .2642 \\
GuidedBackprop ($\mu$) & .2845  & .2851  & .2739  & .2902 \\
GuidedBackprop ($L_2$) & \textbf{.3158}  & .3010  & .2640  & .2642 \\
IntegratedGrads ($\mu$) & .3043  & .2931  & .2860  & \textbf{.2308} \\
IntegratedGrads ($L_2$) & .3098  & \textbf{.3160}  & .2670  & .2800 \\
Activation ($\mu$) & .2781  & .2781  & .2551  & .2551 \\
Activation ($L_2$) & .3111  & .3111  & .3209  & .3209 \\
LIME & .2968  & .3034  & .2888  & .2961 \\
Occlusion & .2898  & .3080  & .2887  & .2656 \\
Shapley & .2908  & .3113  & .2788  & .2592 \\
\bottomrule
\end{tabular}
\caption{
Erasure-based faithfulness results: Average AOPC comprehensiveness and sufficiency scores for different attribution methods on the English split of XNLI. The scores are averaged across all models including different architectures and seeds. Attribution calculations are performed with respect to the top prediction score (TP) and the loss. Different attribution methods perform best for different output mechanisms in terms of comprehensiveness and sufficiency. 
}
\label{erasure-based-results}
\end{table}

Table \ref{erasure-based-results} shows the results of erasure-based faithfulness evaluation (comprehensiveness and sufficiency), for each attribution method. 
In terms of comprehensiveness, 
Saliency and GuidedBackpropagation with $L_2$ aggregation are the most faithful attribution methods when the output is the top prediction score; IntegratedGradients with $L_2$ aggregation is the most faithful one 
when the output is the loss. For sufficiency, InputXGradient with $L_2$ and IntegratedGradients with mean aggregation seem to be the most faithful method for cases when the output is the top prediction score and loss, respectively. Interestingly, most of the results are quite similar and differences between methods are not as large as in the cross-lingual faithfulness evaluation.

\begin{figure}[t!]
    \includegraphics[width=\columnwidth]{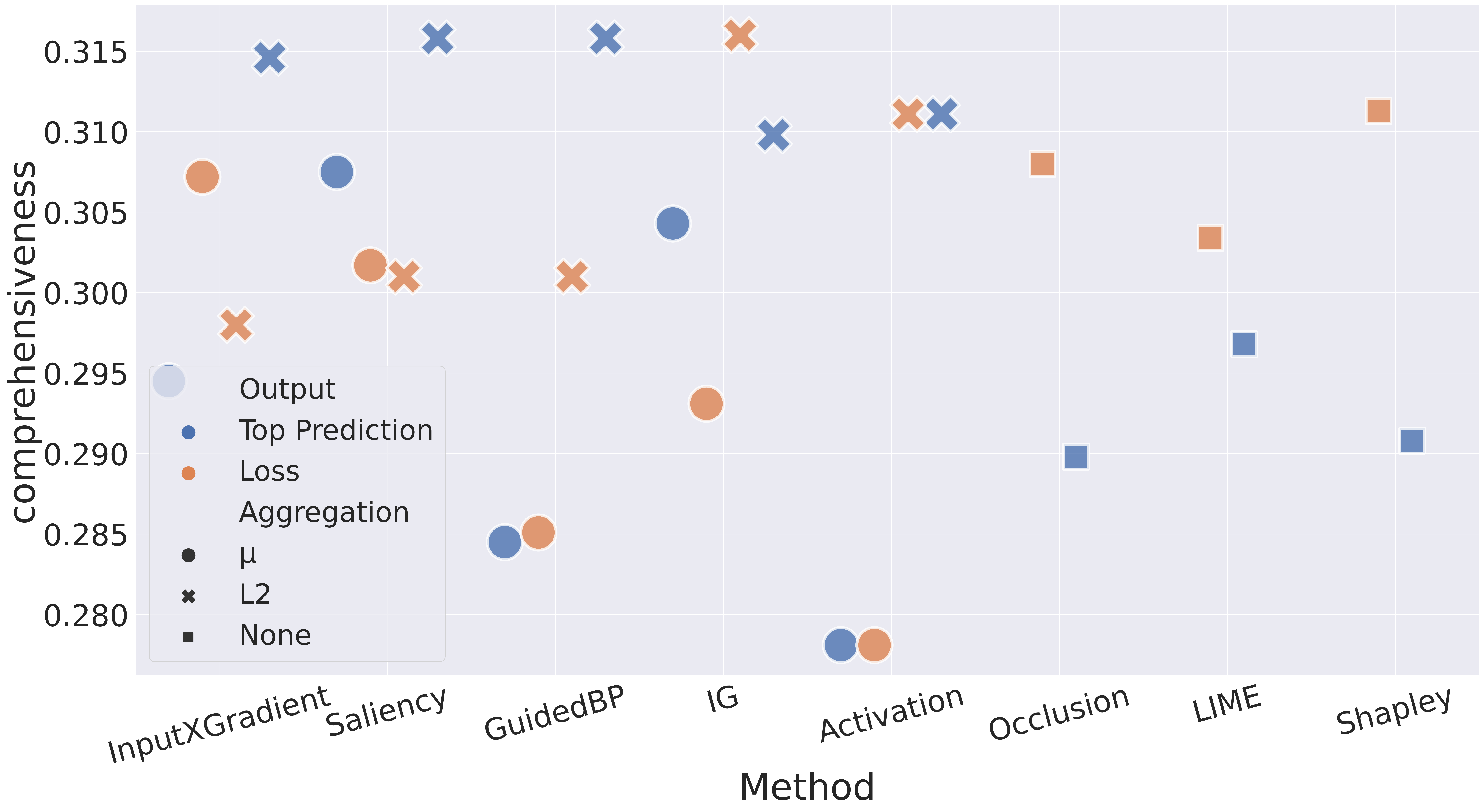}
    \caption{Comparison of comprehensiveness results along output and aggregation dimensions (higher is better). $L_2$ outperforms mean aggregation for most cases and calculations with respect to the loss outperform or are on par with calculations with respect to the top prediction score for non-gradient-based attribution methods.}
    \label{comprehensiveness_results_fig}
\end{figure}

Figure \ref{comprehensiveness_results_fig} shows the effect of aggregation method and output mechanism on comprehensiveness. For all attribution methods, $L_2$  beats mean aggregation except for Saliency and InputXGradient with loss as output. While different output mechanisms are better for different methods, calculating attributions with respect to loss is as good as or slightly better than calculating with respect to the top prediction score for all non-gradient-based methods. 

\begin{figure}[t]
    \includegraphics[width=\columnwidth]{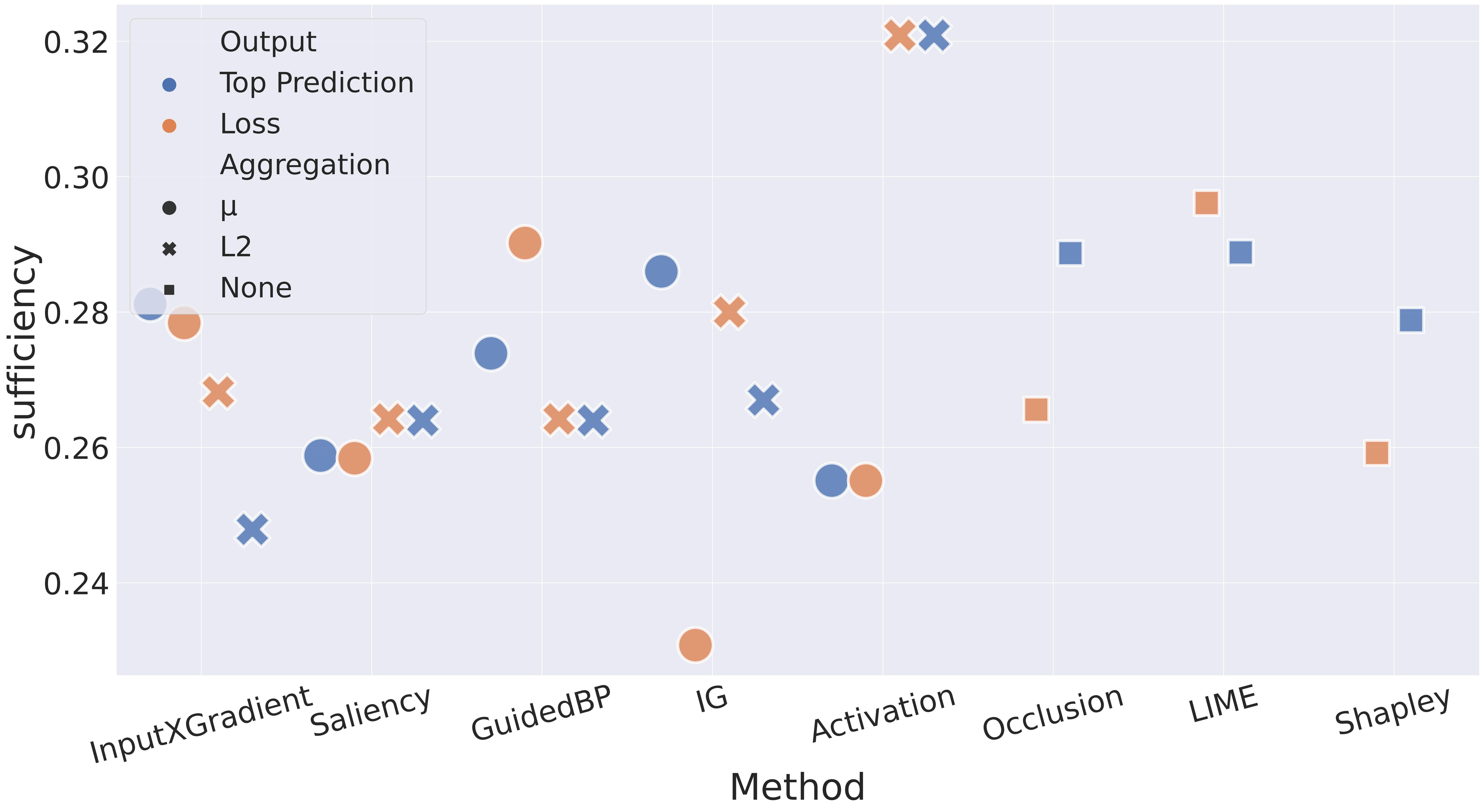}
    \caption{Comparison of sufficiency results along output and aggregation dimensions (lower is better). Different aggregation and different output mechanisms perform better for different attribution methods.}
    \label{suff_results_fig}
\end{figure}

Figure \ref{suff_results_fig} shows the effect of the aggregation method and output mechanism on sufficiency. Unlike comprehensiveness, there is no clear supremacy of one method over another for either aggregation methods or output mechanisms.

\subsubsection{Cross-lingual vs.\ Erasure-based Faithfulness}

The results of cross-lingual faithfulness and erasure-based metrics (comprehensiveness and sufficiency) differ in two main aspects:
\begin{itemize}[leftmargin=10pt,itemsep=2pt,parsep=2pt,topsep=2pt]
\item Perturbation-based methods exhibit more faithful explanations when evaluated by erasure-based metrics  than when evaluated by cross-lingual faithfulness. We interpret this pattern as a result of the OOD issue caused by erasure-based evaluation, which unjustifiably favors perturbation-based attributions. The relative improvement for perturbation-based methods can be attributed to noise due to the OOD perturbations used for calculating comprehensiveness and sufficiency.
\item Erasure-based faithfulness metrics are unable to properly distinguish between different attribution methods, since the differences are dwarfed by the noise introduced by the OOD perturbations. The standard deviation of faithfulness scores across all attribution methods is $0.25$ for cross-lingual faithfulness, but only $0.01$ and $0.02$ for comprehensiveness and sufficiency, respectively.
\end{itemize}

\section{Plausibility}

In this section, we present details about plausibility evaluation and results, and introduce a new dataset containing highlight-based explanations in multiple languages.

\subsection{Plausibility Evaluation}

To evaluate the plausibility of attribution methods, 
we measure agreement with human rationales, following  \citet{atanasova-etal-2020-diagnostic}.  This evaluation measures how much the attribution scores overlap with human annotations by calculating Mean Average Precision (MAP) across a dataset. For each instance in the dataset, Average Precision (AP) is calculated by comparing attribution scores $\boldsymbol{\omega}^{(i)}$ with gold rationales, $\textbf{w}^{(i)}$, where $\boldsymbol{\omega}^{(i)}$ stands for the attribution scores calculated for the dataset instance $\textbf{x}^{(i)}$ and $\textbf{w}^{(i)}$ stands for the sequence of binary labels indicating whether the token is annotated as the rationale. For a dataset  $X = \{ \textbf{x}^{(i)} | i \in [1, M]\}$, the MAP score is defined as:
\begin{equation}
\textrm{MAP}(\omega, X) = \frac{1}{M} \sum_{i \in [1, M]} AP(\textbf{w}^{(i)}, \boldsymbol{\omega}^{(i)}) \end{equation} 

Note that AP is the weighted mean of precisions at each threshold where the weight is the change in recall between two successive thresholds.


\subsection{Plausibility Experiments} \label{sec:plausibility-exp}

\paragraph{Experimental Setup} We use the e-SNLI dataset \citep{Camburu2018eSNLINL} to obtain human annotated highlights. As the classifier, we use a BERT-base model fine-tuned on the SNLI dataset \citep{snli:emnlp2015} with 2 different seeds, as well as the one provided by TextAttack \citep{morris2020textattack}. 

\begin{table}[t]
\centering
\begin{tabular}{lrr}
\toprule
\multirow{2}[3]{*}{\textbf{Method}} & \multicolumn{2}{c}{\textbf{MAP}} \\
\cmidrule(lr){2-3}
 & \textbf{TP} & \textbf{Loss} \\
\midrule
InputXGradient ($\mu$) & .395 & .397 \\
InputXGradient ($L_2$) & .651 & .653 \\
Saliency ($\mu$) & \textbf{.653} & \textbf{.655} \\
Saliency ($L_2$) & \textbf{.653} & \textbf{.655} \\
GuidedBackprop ($\mu$) & .413 & .414 \\
GuidedBackprop ($L_2$) & \textbf{.653} & \textbf{.655} \\
IntegratedGrads ($\mu$) & .473 & .465 \\
IntegratedGrads ($L_2$) & .633 & .599 \\
Activation ($\mu$) & .230 & .230 \\
Activation ($L_2$) & .437 & .437 \\
LIME & .407 & .400 \\
Occlusion & .547 & .476 \\
Shapley & .522 & .460 \\
\bottomrule
\end{tabular}
\caption{
Plausibility results: MAP scores for different attribution methods on the e-SNLI dataset averaged across models. Attribution calculations are performed with respect to the top prediction score (TP) and the loss.  Saliency with both aggregations and GuidedBackprop with $L_2$ aggregation are the best performing methods in both cases.
}
\label{plausibility-results}
\end{table}

\paragraph{Results} Table \ref{plausibility-results} shows GuidedBackprop with $L_2$ aggregation and Saliency with both aggregations are the most plausible methods for both types of output.
Like cross-lingual faithfulness results, gradient-based methods mostly generate more plausible explanations than perturbation-based ones, as in prior work \citep{atanasova-etal-2020-diagnostic}.

\begin{figure}[t]
    \includegraphics[width=\columnwidth]{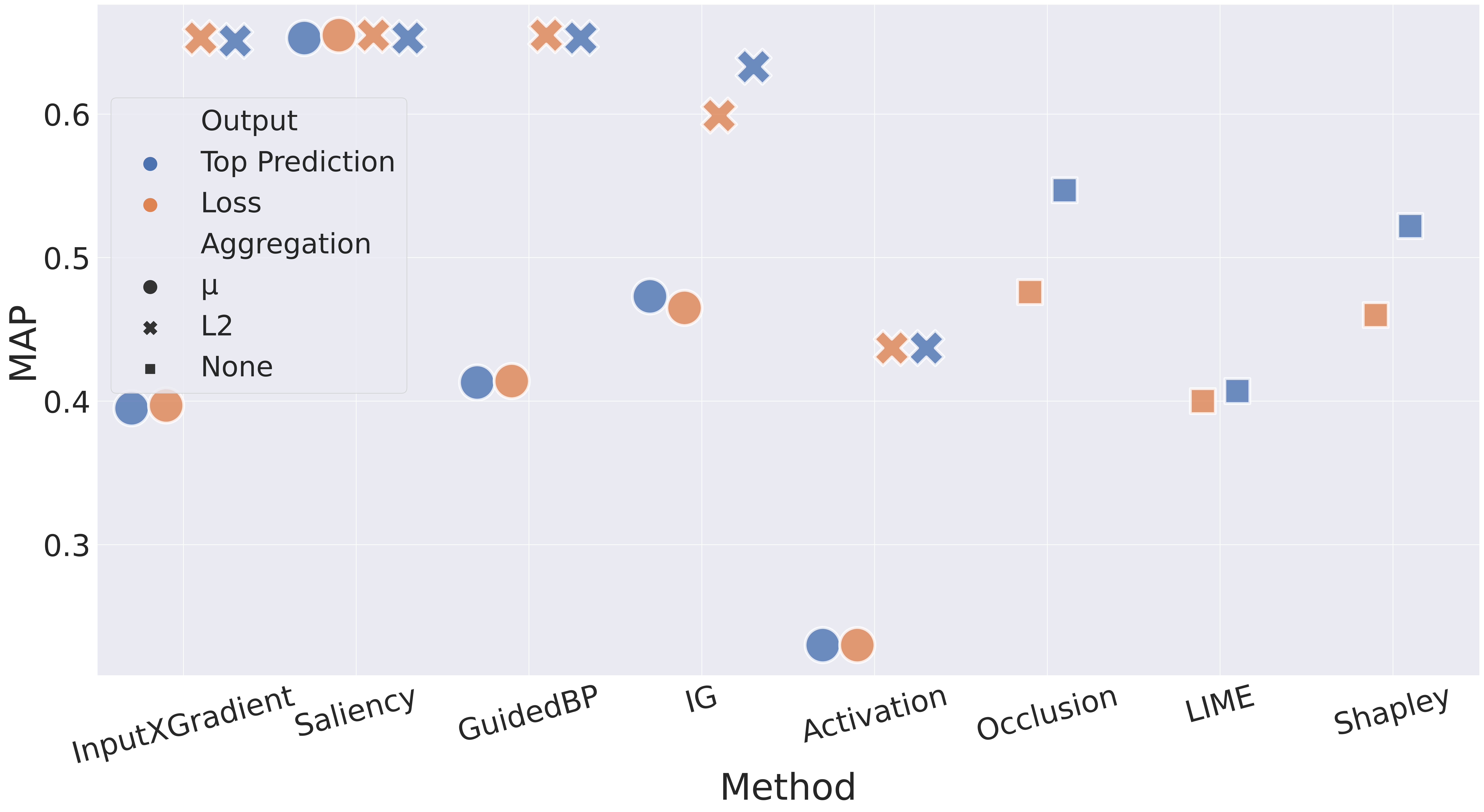}
    \caption{Comparison of plausibility results along output and aggregation dimensions. $L_2$ outperforms mean aggregation for almost all attribution methods and choosing loss as output is mostly the same or slightly better than the top prediction score.}
    \label{human_agreement_results}
\end{figure}

Figure \ref{human_agreement_results} shows the effect of aggregation method and output mechanism on plausibility. In all cases, $L_2$ outperforms mean aggregation by large margins except for Saliency, where the scores for mean aggregation are the same as those for $L_2$ aggregation. Considering that Saliency returns the absolute value, which is analogous to $L_1$ aggregation, the exception in the results makes sense as in the cross-lingual faithfulness results. In almost all cases, calculating attribution scores with respect to loss is the same or slightly better than calculating with respect to the top prediction score. For Integrated Gradients, Occlusion, and LIME, choosing the top prediction score as output outperforms the loss.

\subsection{e-XNLI dataset} Since prior studies for plausibility evaluation are limited to English-only datasets for the NLI task, we augment the multilingual XNLI dataset \citep{conneau2018xnli} with highlight-based explanations by utilizing attribution methods.

First, we compute attribution scores on the English split of the XNLI dataset using an mBERT model fine-tuned on MNLI and Saliency with $L_2$ aggregation and loss as output, which gave the most plausible attribution on e-SNLI (Section~\ref{sec:plausibility-exp}). To extract rationales from the English split, we binarize the attribution scores with respect to the threshold, $0.167$, giving the best F1 score on e-SNLI with the TextAttack model.\footnote{Since there are no human-annotated highlights available for the English split of the XNLI, we first extract rationales by using the  attribution method that perfromed best on a similar dataset having highlights, e-SNLI.} Finally, we project extracted rationales to other languages using \texttt{awesome-align}. 

\looseness=-1
To validate the automatically generated highlights, we follow two approaches. First, we measure the plausibility of the same attribution method used to extract rationales for those languages. This approach investigates whether the aligned rationales are able to follow the same reasoning paths for each language. As Table \ref{exnli-plausibility} shows, the automatically aligned highlights in e-XNLI are similarly plausible explanations for most languages.

\begin{table}
\centering
\begin{tabular}{lr|lr|lr}
\toprule
\textbf{Lang} & \textbf{MAP} & \textbf{Lang} & \textbf{MAP} & \textbf{Lang} & \textbf{MAP} \\
\midrule
ar & 0.663 & es & 0.766 & th & 0.932  \\
bg & 0.701 & fr & 0.739 & tr & 0.665  \\
de & 0.732 & hi & 0.604 & ur & 0.575  \\
el & 0.696 & ru & 0.686 & vi & 0.572  \\
en & 1.0 & sw & 0.58 & zh & 0.543  \\
\bottomrule
\end{tabular}
\caption{Plausibility results: MAP scores measured on the newly introduced e-XNLI dataset (using Saliency with loss as output and $L_2$ aggregation).
}
\label{exnli-plausibility}
\end{table}

Second, we perform a human evaluation on a subset of the created dataset. For four XNLI languages, we sample 10 examples per label (30 total) and request annotators to evaluate the correctness of highlights by following the same procedure carried out in e-SNLI \citep{Camburu2018eSNLINL}. Then, we measure precision, recall, and F1 scores between automatically generated highlights and those manually edited by human annotators. As Table \ref{exnli-human-eval} shows,  automatically generated highlights mostly agree with human reasoning. We present more details about the human evaluation in Appendix \ref{sec:human-evaluation-e-xnli}.

We make the e-XNLI dataset publicly available under MIT license to facilitate research on explainable NLP in a multilingual setting.

\begin{table}[t]
\centering
\begin{tabular}{lrrr}
\toprule
\textbf{Language} & \textbf{Precision} & \textbf{Recall} & \textbf{F1} \\
\midrule
ar &  .64 & .73 & .68 \\
en & .79 & .78 & .79  \\
ru & .93 & .78 & .85  \\
tr & .77 & .71 & .74  \\
\bottomrule
\end{tabular}
\caption{
Human evaluation for a sample of e-XNLI: Precision, recall and F1 scores for four languages.
}
\label{exnli-human-eval}
\end{table}

\section{Limitations}

In this work, we examine a wide range of attribution methods along output and aggregation dimensions. Prior work \citep{Madsen2021EvaluatingTF} shows that faithfulness of attribution methods depends on both tasks and models, but our work is limited to the NLI task while considering different models. 
Despite the importance of NLI for evaluation in NLP \citep{poliak-2020-survey}, our conclusions might not generalize to other tasks. 
In addition, while we experiment with multiple random seeds, our experiments are limited to two architectures: BERT \citep{devlin-etal-2019-bert} and {XLM-R} \citep{Conneau2020UnsupervisedCR}. 

The results of cross-lingual faithfulness experiments are sensitive to language choice as discussed in Section \ref{sec:crosslingual-faithfulness-experiments}, so we present the results calculated with the languages well-represented by multilingual models. The multilingual dataset we provide, e-XNLI, consists of automatically-extracted highlight-based explanations and should be used with caution for future exNLP studies since we only performed a human evaluation on a small subset of the dataset. Especially, training self-explanatory models with this dataset can lead to undesired outcomes, such as poor explanation quality.

\section{Conclusion}

We introduce a novel cross-lingual strategy to evaluate the faithfulness of attribution methods, which eliminates the out-of-distribution input problem of common erasure-based faithfulness evaluations. Then, we perform a comprehensive comparison of different attribution methods having different characteristics in terms of plausibility and faithfulness for the NLI task. The experiments show that there is no one-size-fits-all solution for local post-hoc explanations. Our results highlight that practitioners should choose an attribution method with proper output mechanism and aggregation method according to the property of explanation in question:
\begin{itemize} 

\item For most attribution methods, $L_2$ aggregation and attribution calculation with respect to loss provide more faithful and plausible explanations.

\item Erasure-based faithfulness metrics cannot properly differentiate different attribution methods.

\item Gradient-based attribution methods usually generate more plausible and faithful explanations than perturbation-based methods.

\item To obtain the most plausible explanations, one should choose Guided Backpropagation with $L_2$ and Saliency with either aggregation method, and calculate scores with respect to the loss.

\item To obtain the most faithful explanations, one should choose InputXGradient with $L_2$ regardless of output mechanism.

\end{itemize}

Finally, we present e-XNLI, a multilingual dataset with automatically generated highlight explanations, to facilitate multilingual exNLP studies.

\section*{Acknowledgements}

We would like to thank Adir Rahamim for the ideas on representational similarity experiments of multilingual models, Oleg Serikov for evaluating automatically extracted highlights in the Russian subset of our e-XNLI dataset, and Ramazan Pala for reviews while drafting this paper. 
This research was partly supported by the ISRAEL SCIENCE FOUNDATION (grant No.\ 448/20) and by an Azrieli Foundation Early Career Faculty Fellowship. 

\bibliography{anthology,custom}
\bibliographystyle{acl_natbib}

\clearpage 

\appendix

\section{Cross-lingual performance of multilingual classifiers}
\label{sec:appendix-mbert}

Table~\ref{tab:mbert-scores} shows the average accuracies of the  mBERT and XLM-R models fine-tuned on MNLI for each language in the XNLI dataset. Both models are fine-tuned for 3 epochs with learning rate 2e-5 , total batch size of 256 and 3 different seeds.

\begin{table}[h]
\centering
\begin{tabular}{lll}
\toprule
\textbf{Language} & \textbf{mBERT} & \textbf{XLM-R\textsubscript{base}} \\
\midrule
ar & 0.6574 & 0.7132 \\
bg & 0.6952 & 0.7745 \\
de & 0.7120 & 0.7649 \\
el & 0.6724 & 0.7597 \\
en & \textbf{0.8147} & \textbf{0.8436} \\
es & 0.7504 & 0.7887 \\
fr & 0.7358 & 0.7774 \\
hi & 0.6061 & 0.6959 \\
ru & 0.6906 & 0.7549 \\
sw & 0.5137 & 0.6558 \\
th & 0.5468 & 0.7143 \\
tr & 0.6323 & 0.7269 \\
ur & 0.5856 & 0.6528 \\
vi & 0.7043 & 0.7466 \\
zh & 0.6952 & 0.7318 \\

\bottomrule
\end{tabular}
\caption{\label{tab:mbert-scores}
Accuracies averaged across seeds of the mBERT and XLM-R\textsubscript{base} models fine-tuned on MNLI for each XNLI language.
}
\end{table}

\section{Attribution Methods} \label{sec:appendix-atrributions}

In this work, we focus on a wide range of attribution methods by investigating different combinations of output mechanisms and aggregation methods. We consider two different output options while calculating importance scores per word: (a) top prediction score; (b) loss value calculated when the ground truth label is given. In the following, we refer to the output as $f_{tp}$ when it is the top prediction score and $f_{\mathcal{L}}$ when it is the loss. While some methods inherently return a single score per word, some of them return importance scores for each dimension of the corresponding word vector. Since we want to obtain a single score per word, those scores need to be aggregated. We investigate $L_2$ and mean aggregations separately.

\paragraph{Implementation Details} We build our framework upon the Captum library \citep{kokhlikyan2020captum} to use existing implementations of many attribution methods. We use the HuggingFace transformers \citep{wolf-etal-2020-transformers} and datasets \citep{lhoest-etal-2021-datasets} libraries to access pretrained models and datasets. Also, we rely upon Scikit-learn \citep{scikit-learn} for evaluation scores such as Average Precision (AP) and Spearman Correlation.

\subsection{Saliency}

Saliency \citep{Simonyan2014DeepIC} calculates attribution scores by calculating the absolute value of the gradients with respect to inputs. More formally, let $u_j$ be the embedding for word $x_j$ of $\textbf{x}^{(i)}$, the $i$-th instance of any dataset. Then the attribution score per each dimension of the embedding is defined as \begin{equation} \big| \nabla_{u_{jk}} f(\textbf{x}^{(i)}) \big| \end{equation}

We obtain an attribution score per word, $\omega^{(i)}_{x_j}$, by aggregating scores across each word embedding. Using mean aggregation, it is defined as follows:
\begin{equation} \omega^{(i)}_{x_j} = \frac{1}{d} \sum_{k=0}^d \big| \nabla_{u_{jk}} f(\textbf{x}^{(i)}) \big| \end{equation} where $d$ is the number of dimensions for the word embedding.
Similarly, using $L_2$ aggregation, we obtain
\begin{equation} \omega^{(i)}_{x_j} = \sqrt{\sum_{k=0}^d \big| \nabla_{u_{jk}} f(\textbf{x}^{(i)}) \big|^2} \end{equation}

\subsection{InputXGradient}

InputXGradient \citep{Shrikumar2016NotJA} calculates attribution scores by multiplying the input with the gradients with respect to the input. More formally, the attribution score per each dimension is defined as 
\begin{equation} \nabla_{u_{jk}} f(\textbf{x}^{(i)}) u_{jk} \end{equation}

We obtain attribution scores per word in the same way as Saliency using mean/$L_2$ aggregations.

\subsection{Guided Backpropagation}

Guided Backpropagation \citep{Springenberg2015StrivingFS} produces attribution scores by calculating gradients with respect to the input. Different from other methods, it overrides the gradient of the ReLU activation so that only positive gradients pass through. We obtain attribution scores per word using $L_2$ and mean aggregations as in the previously described methods.

\subsection{Integrated Gradients}

Integrated Gradients \citep{integrated-grads} produces attribution scores by summing gradients along each dimension from some baseline input to a given input. The attribution score per each dimension is defined as
\begin{equation} u^{(i)}_{jk} - \overline{u}^{(i)}_{jk} \times   \sum^{m}_{l=1} \frac{\partial f(\overline{u}^{(i)}_{jk} + \frac{l}{m} \times (u^{(i)}_{jk} - \overline{u}^{(i)}_{jk}) ) }{\partial u^{(i)}_{jk}} \times \frac{1}{m} \end{equation}
where $m$ is the number of steps for a Riemannian approximation of the path integral and $\overline{u}^{(i)}_{j}$ is the baseline input. We use the word embedding of the [PAD] token as the baseline input for each word except for [SEP] and [CLS] tokens \citep{sajjad2021interpretiontutorial}. We obtain attribution scores per word using $L_2$ and mean aggregations as in the previous methods.

Higher values of $m$ would produce a better approximation, but also make attribution calculation computationally expensive. We need to find a sweet spot between approximation and computational resources. For plausibility experiments, we select $m$ according to validation performance based on  MAP scores. Among $\{50, 75, 100\}$, we choose $m = 100$ for mean aggregation on calculations with respect to top prediction and $m = 50$ for all other combinations of aggregation methods and output mechanisms. For cross-lingual faithfulness experiments, we select $m$ according to the evaluation on the validation set based on the Spearman correlation coefficient values. Among \{50,75,100\}, we choose $m=100$ for all calculations with XLM-R\textsubscript{base} and mBERT except calculations involving mean aggregation on mBERT, for which we choose $m = 75$. For erasure-based faithfulness experiments, we use the same values of $m$ for the sake of a fair comparison.

\subsection{LIME}

LIME \citep{Ribeiro2016WhySI} produces attribution scores by training a surrogate linear model using the points around the input created by perturbing the input and output of perturbations from the original model. A random subset of the input is replaced by a baseline value to create perturbations. We use the word embedding of the [PAD] token as the baseline value (as in Integrated Gradients). Since we create the perturbations by replacing whole word vectors, we obtain a single score per word, which eliminates the need for aggregation. We use 50 samples for training the surrogate model as the default value for the LIME implementation in Captum.

\subsection{Occlusion}

Occlusion \citep{Zeiler2014VisualizingAU} produces attribution scores by calculating differences in the output after replacing the input with baseline values over a sliding window. We select the shape of the sliding window so that it occludes only the embedding of one word at a time, and we use the word embedding of the [PAD] token as a baseline value (as in Integrated Gradients and LIME). Since we create the perturbations by replacing whole word vectors, we obtain a single score per word.

\subsection{Shapley Value Sampling}

In Shapley Value Sampling \citep{trumbelj2010AnEE}, we take a random permutation of input, which is word embeddings of input sequence in our case, and add them one by one to a given baseline, embedding vector for [PAD] token in our case, to produce attribution score by calculating the difference in the output. The scores are averaged across several samples. We choose the feature group so that one score corresponds to a single word, which eliminates the need for aggregation. We take 25 samples for calculating attributions as the default value for Shapley Value Sampling implementation in Captum.

\subsection{Activation}

Layer Activation \citep{Karpathy2015VisualizingAU} produces attribution scores by getting the activations in the output of the specified layer. We select the embedding layer for this purpose, which yields an attribution score per each dimension of the embedding equal to $u_{jk}$. Then, we obtain attribution scores per word using $L_2$ and mean aggregations as in other methods.

\section{Representational Similarity of Translation Pairs}
\label{sec:input-similarity}

Our cross-lingual faithfulness strategy relies on the assumption that translation pairs constitute similar inputs for a multilingual model. To test our assumption, we create a setup comparing representational similarities of inputs. First, we take premise-hypothesis pairs and their translations from XNLI dataset for the selected language pair. We encode each pair by obtaining the last hidden state representations before the classifier head. We take $n$ representations from the source language and the corresponding representations from the target language to create source and target batch pairs, namely $(b_s, b_t)$. Then, we create $k$ random batches, $b_i$, by selecting $n$ representations among target representations for each one and we measure the CKA similarity \citep{pmlr-v97-kornblith19a} of representation batch in the source language with each batch of representation in the target language. For the sake of our assumption, we expect matching representation batches to be more similar than any batch pairs. For each batch in the source langauge, we test whether the CKA similarity measure assigns the highest similarity to the matching batches or not and compute the accuracy over batches.

We use 5000 examples from the test split of the XNLI dataset by selecting $n = 8$ and $k=10$.

\begin{figure}[t]
    \includegraphics[width=\columnwidth]{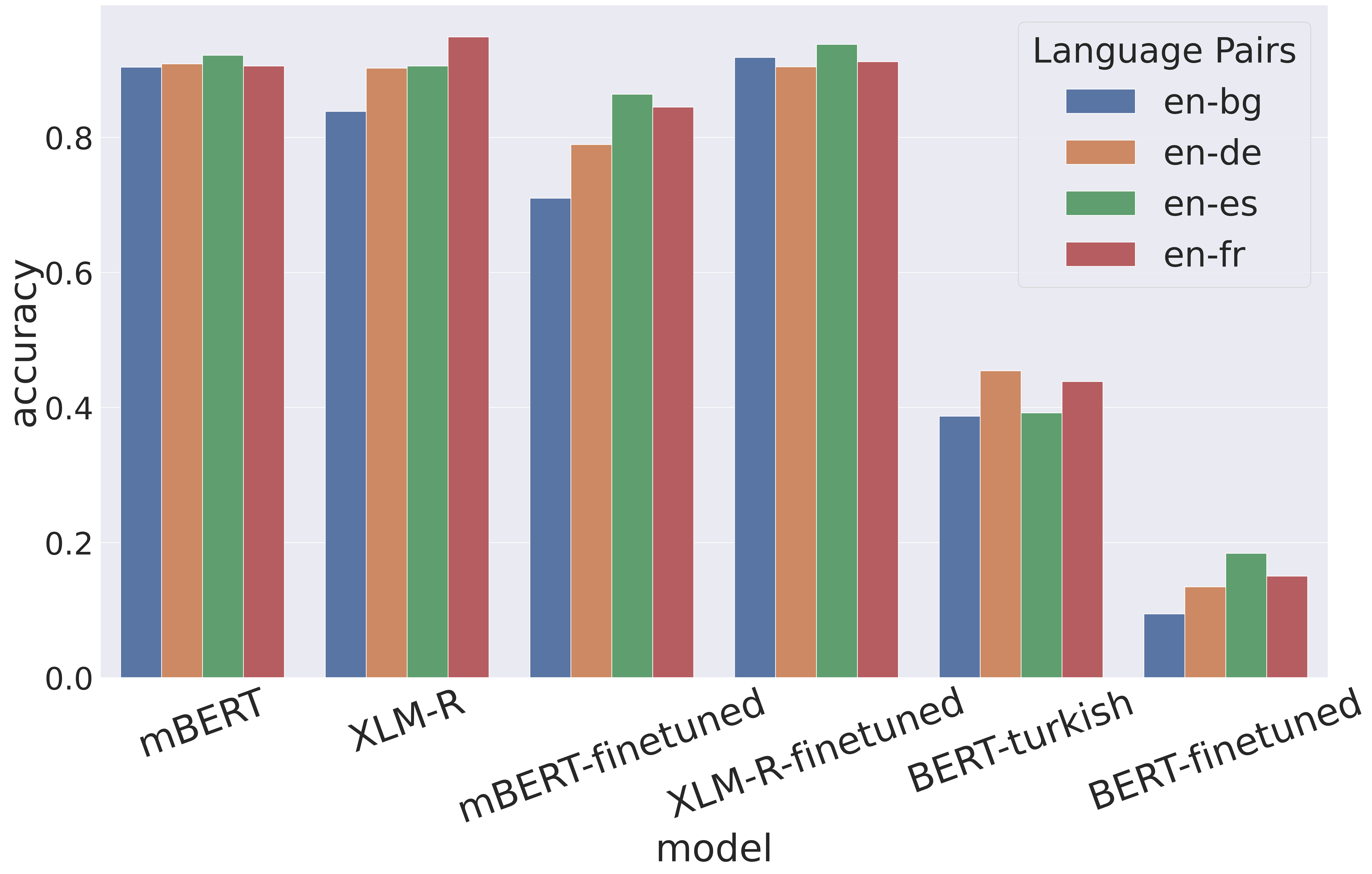}
    \caption{Accuracies for CKA similarity analysis for different models. XLM-R-finetuned and mBERT-finetuned results are averaged across models fine-tuned with different seeds for each.}
    \label{input_similarity_results}
\end{figure}

Figure \ref{input_similarity_results} shows accuracies for different models. We perform our similarity analysis on multilingual models, which are vanilla mBERT, mBERT and XLM-R\textsubscript{base} models fine-tuned on MNLI and used in our faithfulness experiments, and monolingual models, which are BERT\textsubscript{base} fine-tuned on MNLI and a Turkish BERT \cite{stefan_schweter_2020_3770924}. The results show that translation pairs form similar inputs for multilingual models compared to monolingual models regardless of being fine-tuned. While the accuracies of fine-tuned mBERT are lower than standard mBERT, it differs among language pairs for XLM-R\textsubscript{base} case. Although monolingual representations of translation pairs lead to the lowest accuracies as expected, higher accuracies of Turkish BERT, which is pre-trained on a completely unrelated language, compared to fine-tuned English BERT\textsubscript{base} need further investigation.

\section{Ablation Study on Cross-lingual Faithfulness}
\label{sec:crosslingual-faithfulness-ablation}

To investigate the effect of word alignments, we run our cross-lingual faithfulness evaluation framework with random word alignments for a set of attribution methods and compare the results with the ones obtained with \texttt{awesome-align} \citep{dou-neubig-2021-word}. To obtain random alignments between translation pairs, we modify the IterMax algorithm, which \citet{dou-neubig-2021-word} proposed as a baseline method, by replacing the similarity matrix with a random matrix. We perform both types of evaluations with one of the mBERT models we fine-tuned. 

\begin{figure}[t]
    \includegraphics[width=\columnwidth]{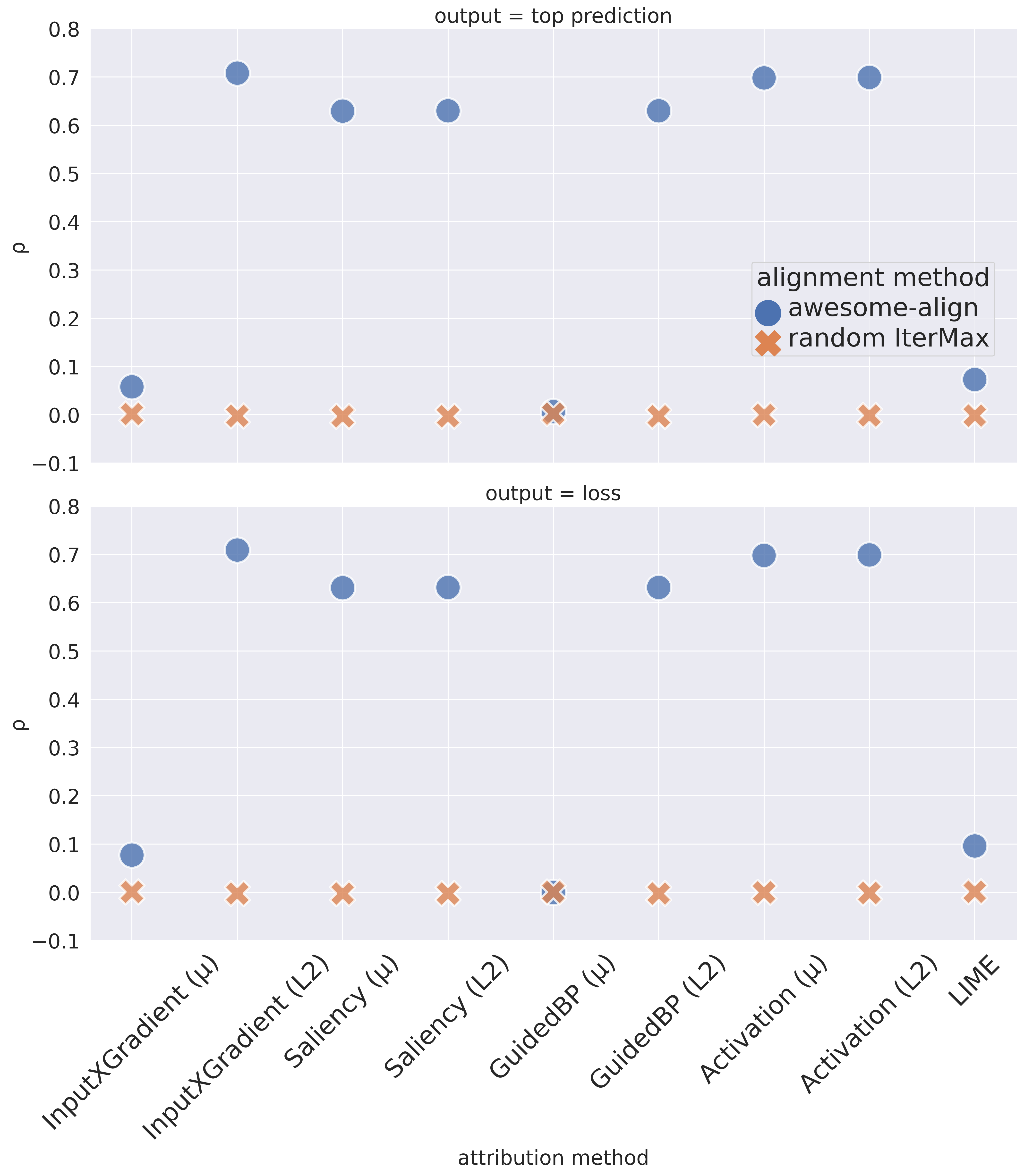}
    \caption{Comparison of cross-lingual faithfulness scores that are calculated with \texttt{awesome-align} and random word alignments for different attribution methods}
    \label{alignment_ablation_results}
\end{figure}

Figure \ref{alignment_ablation_results} shows the comparison of awesome-align with random word alignments. While using \texttt{awesome-align} provides comparable scores across attribution methods, random alignments lead to near-zero correlations ($\rho$). Thus we empirically show that word alignment forms a significant part of our method.

\section{Cross-lingual Faithfulness Results per Architecture}
\label{sec:crosslingual-faithfulness-results-per-arch}

Table \ref{tab:crosslingualfaithfulness-results-per-arch} shows cross-lingual faithfulness results for each architecture, mBERT and XLM-R\textsubscript{base}, separately.

\begin{table*}[t]
\centering
\begin{tabular}{lrrrr}
\toprule
\multirow{3}[3]{*}{\textbf{Method}} & \multicolumn{4}{c}{\textbf{$\rho$}} \\
\cmidrule(lr){2-5}
 &  \multicolumn{2}{c}{\textbf{mBERT}} & \multicolumn{2}{c}{\textbf{XLM-R\textsubscript{base}}} \\
 \cmidrule(lr){2-5}
 & \textbf{TP} & \textbf{Loss} & \textbf{TP} & \textbf{Loss} \\
 \midrule
InputXGradient ($\mu$) & .0562 $\pm$ .002 & .0758 $\pm$ .002 & .0615 $\pm$ .001 & .0754 $\pm$ .003 \\
InputXGradient ($L_2$) & .7067 $\pm$ .001 & .7078 $\pm$ .001 & .7336 $\pm$ .003 & .7338 $\pm$ .003 \\
Saliency ($\mu$) & .6269 $\pm$ .003 & .6283 $\pm$ .003 & .5082 $\pm$ .001 & .5078 $\pm$ .002 \\
Saliency ($L_2$) & .6276 $\pm$ .003 & .6290 $\pm$ .003 & .5053 $\pm$ .001 & .5050 $\pm$ .002 \\
GuidedBackprop ($\mu$) & .0024 $\pm$ .003 & -.0000 $\pm$ .001 & .0028 $\pm$ .000 & .0041 $\pm$ .001 \\
GuidedBackprop ($L_2$) & .6276 $\pm$ .003 & .6290 $\pm$ .003 & .5053 $\pm$ .001 & .5050 $\pm$ .002 \\
IntegratedGrads ($\mu$) & .1860 $\pm$ .008 & .2680 $\pm$ .007 & .1897 $\pm$ .021 & .2198 $\pm$ .008 \\
IntegratedGrads ($L_2$) & .5910 $\pm$ .009 & .5302 $\pm$ .005 & .6279 $\pm$ .018 & .5970 $\pm$ .017 \\
Activation ($\mu$) & .6974 $\pm$ .001 & .6974 $\pm$ .001 & .4130 $\pm$ .001 & .4130 $\pm$ .001 \\
Activation ($L_2$) & .6992 $\pm$ .000 & .6992 $\pm$ .000 & .6938 $\pm$ .000 & .6938 $\pm$ .000 \\
LIME & .0659 $\pm$ .014 & .0934 $\pm$ .006 & .0182 $\pm$ .009 & .0420 $\pm$ .008 \\
Occlusion & .2281 $\pm$ .007 & .3132 $\pm$ .005 & .0680 $\pm$ .028 & .0966 $\pm$ .007 \\
Shapley & .3734 $\pm$ .049 & .4058 $\pm$ .040 & .0833 $\pm$ .016 & .1426 $\pm$ .033 \\
\bottomrule
 \end{tabular}
\caption{
Cross-lingual faithfulness results: Scores are measured for different attribution methods on the XNLI dataset and averaged across models trained with different seeds for each architecture. Attribution calculations are performed with respect to the top prediction (TP) score and the loss.
}
 \label{tab:crosslingualfaithfulness-results-per-arch}
 \end{table*}

\section{Erasure-based Faithfulness Results per Architecture}
\label{sec:eraser-faithfulness-results-per-arch}

Table \ref{tab:comprehensiveness-results-per-arch} and Table \ref{tab:sufficiency-results-per-arch} show comprehensiveness and sufficiency scores for each architecture, mBERT and XLM-R\textsubscript{base}, separately.

\begin{table*}[t]
\centering
\begin{tabular}{lrrrr}
\toprule
\multirow{3}[3]{*}{\textbf{Method}} & \multicolumn{4}{c}{\textbf{comprehensiveness} $\uparrow$} \\
\cmidrule(lr){2-5}
 &  \multicolumn{2}{c}{\textbf{mBERT}} & \multicolumn{2}{c}{\textbf{XLM-R\textsubscript{base}}} \\
 \cmidrule(lr){2-5}
 & \textbf{TP} & \textbf{Loss} & \textbf{TP} & \textbf{Loss} \\
 \midrule
InputXGradient ($\mu$) & .2658 $\pm$ .016 & .2959 $\pm$ .012 & .3232 $\pm$ .011 & .3186 $\pm$ .031 \\
InputXGradient ($L_2$) & .3136 $\pm$ .011 & .3080 $\pm$ .005 & .3155 $\pm$ .021 & .2880 $\pm$ .005 \\
Saliency ($\mu$) & .3009 $\pm$ .018 & .2891 $\pm$ .036 & .3141 $\pm$ .008 & .3142 $\pm$ .005 \\
Saliency ($L_2$) & .3128 $\pm$ .018 & .2896 $\pm$ .037 & .3188 $\pm$ .009 & .3123 $\pm$ .004 \\
GuidedBackprop ($\mu$) & .2709 $\pm$ .002 & .2514 $\pm$ .039 & .2981 $\pm$ .023 & .3187 $\pm$ .015 \\
GuidedBackprop ($L_2$) & .3128 $\pm$ .018 & .2896 $\pm$ .037 & .3188 $\pm$ .009 & .3123 $\pm$ .004 \\
IntegratedGrads ($\mu$) & .2557 $\pm$ .033 & .2618 $\pm$ .010 & .3529 $\pm$ .021 & .3244 $\pm$ .018 \\
IntegratedGrads ($L_2$) & .2989 $\pm$ .004 & .2969 $\pm$ .014 & .3208 $\pm$ .028 & .3350 $\pm$ .031 \\
Activation ($\mu$) & .2504 $\pm$ .009 & .2504 $\pm$ .009 & .3057 $\pm$ .006 & .3057 $\pm$ .006 \\
Activation ($L_2$) & .2940 $\pm$ .010 & .2940 $\pm$ .010 & .3282 $\pm$ .017 & .3282 $\pm$ .017 \\
LIME & .2733 $\pm$ .026 & .2657 $\pm$ .016 & .3203 $\pm$ .028 & .3412 $\pm$ .024 \\
Occlusion & .2727 $\pm$ .034 & .3101 $\pm$ .014 & .3068 $\pm$ .029 & .3060 $\pm$ .016 \\
Shapley & .2660 $\pm$ .032 & .3123 $\pm$ .007 & .3157 $\pm$ .019 & .3103 $\pm$ .023 \\
\bottomrule
 \end{tabular}
\caption{
Comprehensiveness scores per architecture on the English split of XNLI dataset: Scores are measured for different attribution methods on the XNLI dataset and averaged across models trained with different seeds for each architecture. Attribution calculations are performed with respect to the top prediction (TP) score and the loss.
}
 \label{tab:comprehensiveness-results-per-arch}
 \end{table*}
 
 \begin{table*}[t]
\centering
\begin{tabular}{lrrrr}
\toprule
\multirow{3}[3]{*}{\textbf{Method}} & \multicolumn{4}{c}{\textbf{sufficiency} $\downarrow$} \\
\cmidrule(lr){2-5}
 &  \multicolumn{2}{c}{\textbf{mBERT}} & \multicolumn{2}{c}{\textbf{XLM-R\textsubscript{base}}} \\
 \cmidrule(lr){2-5}
 & \textbf{TP} & \textbf{Loss} & \textbf{TP} & \textbf{Loss} \\
 \midrule
InputXGradient ($\mu$) & .2812 $\pm$ .013 & .2716 $\pm$ .021 & .2812 $\pm$ .007 & .2852 $\pm$ .014 \\
InputXGradient ($L_2$) & .2616 $\pm$ .043 & .2684 $\pm$ .027 & .2342 $\pm$ .023 & .2681 $\pm$ .026 \\
Saliency ($\mu$) & .2451 $\pm$ .028 & .2613 $\pm$ .029 & .2724 $\pm$ .011 & .2555 $\pm$ .004 \\
Saliency ($L_2$) & .2477 $\pm$ .022 & .2629 $\pm$ .027 & .2804 $\pm$ .008 & .2654 $\pm$ .007 \\
GuidedBackprop ($\mu$) & .2637 $\pm$ .031 & .2913 $\pm$ .007 & .2841 $\pm$ .032 & .2891 $\pm$ .023 \\
GuidedBackprop ($L_2$) & .2477 $\pm$ .022 & .2629 $\pm$ .027 & .2804 $\pm$ .008 & .2654 $\pm$ .007 \\
IntegratedGrads ($\mu$) & .2985 $\pm$ .008 & .2471 $\pm$ .024 & .2734 $\pm$ .020 & .2145 $\pm$ .011 \\
IntegratedGrads ($L_2$) & .2784 $\pm$ .010 & .2788 $\pm$ .021 & .2556 $\pm$ .006 & .2812 $\pm$ .025 \\
Activation ($\mu$) & .2024 $\pm$ .017 & .2024 $\pm$ .017 & .3079 $\pm$ .010 & .3079 $\pm$ .010 \\
Activation ($L_2$) & .3340 $\pm$ .003 & .3340 $\pm$ .003 & .3078 $\pm$ .005 & .3078 $\pm$ .005 \\
LIME & .2610 $\pm$ .005 & .2610 $\pm$ .012 & .3167 $\pm$ .048 & .3311 $\pm$ .005 \\
Occlusion & .2820 $\pm$ .006 & .2475 $\pm$ .008 & .2955 $\pm$ .015 & .2837 $\pm$ .006 \\
Shapley & .2538 $\pm$ .008 & .1967 $\pm$ .008 & .3037 $\pm$ .035 & .3218 $\pm$ .013 \\
\bottomrule
 \end{tabular}
\caption{
Sufficiency scores per architecture on the English split of XNLI dataset: Scores are measured for different attribution methods on the XNLI dataset and averaged across models trained with different seeds for each architecture. Attribution calculations are performed with respect to the top prediction (TP) score and the loss.
}
 \label{tab:sufficiency-results-per-arch}
 \end{table*}

\section{Cross-lingual Faithfulness Results per Language}
\label{sec:crosslingual-faithfulness-results-per-lang}

Our cross-lingual faithfulness evaluation averages correlations across languages. For completeness, we provide in Tables~\ref{crosslingualfaithfulness-results-bulgarian}--\ref{crosslingualfaithfulness-results-urdu} the results of cross-lingual faithfulness evaluation per language.


\begin{table*}[t]
\centering
\begin{tabular}{lrrrr}
\toprule
\multirow{3}[3]{*}{\textbf{Method}} & \multicolumn{4}{c}{\textbf{$\rho$}} \\
\cmidrule(lr){2-5}
 &  \multicolumn{2}{c}{\textbf{mBERT}} & \multicolumn{2}{c}{\textbf{XLM-R\textsubscript{base}}} \\
 \cmidrule(lr){2-5}
 & \textbf{TP} & \textbf{Loss} & \textbf{TP} & \textbf{Loss} \\
 \midrule
InputXGradient ($\mu$) & .0302 $\pm$ .001 & .0406 $\pm$ .001 & .0506 $\pm$ .003 & .0661 $\pm$ .002 \\
InputXGradient ($L_2$) & .6731 $\pm$ .001 & .6741 $\pm$ .002 & .7188 $\pm$ .003 & .7189 $\pm$ .003 \\
Saliency ($\mu$) & .5778 $\pm$ .004 & .5793 $\pm$ .004 & .4954 $\pm$ .000 & .4951 $\pm$ .002 \\
Saliency ($L_2$) & .5787 $\pm$ .004 & .5803 $\pm$ .004 & .4935 $\pm$ .000 & .4930 $\pm$ .001 \\
GuidedBackprop ($\mu$) & .0015 $\pm$ .001 & -.0045 $\pm$ .003 & .0020 $\pm$ .001 & .0023 $\pm$ .004 \\
GuidedBackprop ($L_2$) & .5787 $\pm$ .004 & .5803 $\pm$ .004 & .4935 $\pm$ .000 & .4930 $\pm$ .001 \\
IntegratedGrads ($\mu$) & .1248 $\pm$ .005 & .2003 $\pm$ .002 & .1768 $\pm$ .021 & .2037 $\pm$ .009 \\
IntegratedGrads ($L_2$) & .5287 $\pm$ .014 & .4585 $\pm$ .011 & .6165 $\pm$ .016 & .5827 $\pm$ .016 \\
Activation ($\mu$) & .6080 $\pm$ .001 & .6080 $\pm$ .001 & .3824 $\pm$ .001 & .3824 $\pm$ .001 \\
Activation ($L_2$) & .6653 $\pm$ .000 & .6653 $\pm$ .000 & .6825 $\pm$ .000 & .6825 $\pm$ .000 \\
LIME & .0561 $\pm$ .015 & .0803 $\pm$ .008 & .0115 $\pm$ .007 & .0359 $\pm$ .009 \\
Occlusion & .1635 $\pm$ .008 & .2395 $\pm$ .004 & .0509 $\pm$ .021 & .0754 $\pm$ .004 \\
Shapley & .3348 $\pm$ .055 & .3639 $\pm$ .044 & .0649 $\pm$ .016 & .1165 $\pm$ .032 \\
\bottomrule
 \end{tabular}
\caption{
Cross-lingual faithfulness results for the Bulgarian split of XNLI dataset: Scores are measured for different attribution methods on the XNLI dataset and averaged across models trained with different seeds for each architecture. Attribution calculations are performed with respect to the top prediction (TP) score and the loss.
}
 \label{crosslingualfaithfulness-results-bulgarian}
 \end{table*}


\begin{table*}[t]
\centering
\begin{tabular}{lrrrr}
\toprule
\multirow{3}[3]{*}{\textbf{Method}} & \multicolumn{4}{c}{\textbf{$\rho$}} \\
\cmidrule(lr){2-5}
 &  \multicolumn{2}{c}{\textbf{mBERT}} & \multicolumn{2}{c}{\textbf{XLM-R\textsubscript{base}}} \\
 \cmidrule(lr){2-5}
 & \textbf{TP} & \textbf{Loss} & \textbf{TP} & \textbf{Loss} \\
 \midrule
InputXGradient ($\mu$) & .0493 $\pm$ .003 & .0717 $\pm$ .004 & .0621 $\pm$ .001 & .0752 $\pm$ .003 \\
InputXGradient ($L_2$) & .7052 $\pm$ .001 & .7067 $\pm$ .001 & .7321 $\pm$ .003 & .7329 $\pm$ .003 \\
Saliency ($\mu$) & .6152 $\pm$ .003 & .6168 $\pm$ .003 & .4936 $\pm$ .002 & .4929 $\pm$ .004 \\
Saliency ($L_2$) & .6159 $\pm$ .003 & .6175 $\pm$ .003 & .4906 $\pm$ .002 & .4900 $\pm$ .003 \\
GuidedBackprop ($\mu$) & .0041 $\pm$ .005 & .0008 $\pm$ .002 & .0030 $\pm$ .001 & .0010 $\pm$ .001 \\
GuidedBackprop ($L_2$) & .6159 $\pm$ .003 & .6175 $\pm$ .003 & .4906 $\pm$ .002 & .4900 $\pm$ .003 \\
IntegratedGrads ($\mu$) & .1919 $\pm$ .011 & .2788 $\pm$ .011 & .1814 $\pm$ .019 & .2219 $\pm$ .008 \\
IntegratedGrads ($L_2$) & .5935 $\pm$ .007 & .5361 $\pm$ .005 & .6245 $\pm$ .018 & .5930 $\pm$ .017 \\
Activation ($\mu$) & .6960 $\pm$ .000 & .6960 $\pm$ .000 & .4115 $\pm$ .001 & .4115 $\pm$ .001 \\
Activation ($L_2$) & .7012 $\pm$ .000 & .7012 $\pm$ .000 & .6934 $\pm$ .000 & .6934 $\pm$ .000 \\
LIME & .0692 $\pm$ .012 & .0955 $\pm$ .005 & .0186 $\pm$ .009 & .0399 $\pm$ .008 \\
Occlusion & .2226 $\pm$ .006 & .3117 $\pm$ .005 & .0695 $\pm$ .028 & .0978 $\pm$ .005 \\
Shapley & .3843 $\pm$ .046 & .4145 $\pm$ .036 & .0840 $\pm$ .017 & .1381 $\pm$ .030 \\
\bottomrule
 \end{tabular}
\caption{
Cross-lingual faithfulness results for the German split of XNLI dataset: Scores are measured for different attribution methods on the XNLI dataset and averaged across models trained with different seeds for each architecture. Attribution calculations are performed with respect to the top prediction (TP) score and the loss.
}
 \label{crosslingualfaithfulness-results-german}
 \end{table*}


\begin{table*}[t]
\centering
\begin{tabular}{lrrrr}
\toprule
\multirow{3}[3]{*}{\textbf{Method}} & \multicolumn{4}{c}{\textbf{$\rho$}} \\
\cmidrule(lr){2-5}
 &  \multicolumn{2}{c}{\textbf{mBERT}} & \multicolumn{2}{c}{\textbf{XLM-R\textsubscript{base}}} \\
 \cmidrule(lr){2-5}
 & \textbf{TP} & \textbf{Loss} & \textbf{TP} & \textbf{Loss} \\
 \midrule
InputXGradient ($\mu$) & .0747 $\pm$ .001 & .1009 $\pm$ .003 & .0726 $\pm$ .003 & .0856 $\pm$ .005 \\
InputXGradient ($L_2$) & .7179 $\pm$ .001 & .7186 $\pm$ .001 & .7460 $\pm$ .003 & .7461 $\pm$ .003 \\
Saliency ($\mu$) & .6576 $\pm$ .002 & .6592 $\pm$ .002 & .5205 $\pm$ .001 & .5205 $\pm$ .002 \\
Saliency ($L_2$) & .6582 $\pm$ .002 & .6598 $\pm$ .002 & .5170 $\pm$ .001 & .5172 $\pm$ .001 \\
GuidedBackprop ($\mu$) & .0004 $\pm$ .003 & .0026 $\pm$ .003 & .0035 $\pm$ .003 & .0049 $\pm$ .002 \\
GuidedBackprop ($L_2$) & .6582 $\pm$ .002 & .6598 $\pm$ .002 & .5170 $\pm$ .001 & .5172 $\pm$ .001 \\
IntegratedGrads ($\mu$) & .2189 $\pm$ .008 & .3041 $\pm$ .005 & .2037 $\pm$ .020 & .2374 $\pm$ .006 \\
IntegratedGrads ($L_2$) & .6145 $\pm$ .010 & .5584 $\pm$ .006 & .6384 $\pm$ .018 & .6083 $\pm$ .018 \\
Activation ($\mu$) & .7521 $\pm$ .001 & .7521 $\pm$ .001 & .4311 $\pm$ .001 & .4311 $\pm$ .001 \\
Activation ($L_2$) & .7071 $\pm$ .000 & .7071 $\pm$ .000 & .7135 $\pm$ .000 & .7135 $\pm$ .000 \\
LIME & .0716 $\pm$ .013 & .1032 $\pm$ .009 & .0259 $\pm$ .012 & .0500 $\pm$ .006 \\
Occlusion & .2693 $\pm$ .008 & .3594 $\pm$ .006 & .0839 $\pm$ .035 & .1202 $\pm$ .014 \\
Shapley & .3928 $\pm$ .047 & .4238 $\pm$ .039 & .1041 $\pm$ .022 & .1716 $\pm$ .035 \\
\bottomrule
 \end{tabular}
\caption{
Cross-lingual faithfulness results for the Spanish split of XNLI dataset: Scores are measured for different attribution methods on the XNLI dataset and averaged across models trained with different seeds for each architecture. Attribution calculations are performed with respect to the top prediction (TP) score and the loss.
}
 \label{crosslingualfaithfulness-results-spanish}
 \end{table*}

 
\begin{table*}[t]
\centering
\begin{tabular}{lrrrr}
\toprule
\multirow{3}[3]{*}{\textbf{Method}} & \multicolumn{4}{c}{\textbf{$\rho$}} \\
\cmidrule(lr){2-5}
 &  \multicolumn{2}{c}{\textbf{mBERT}} & \multicolumn{2}{c}{\textbf{XLM-R\textsubscript{base}}} \\
 \cmidrule(lr){2-5}
 & \textbf{TP} & \textbf{Loss} & \textbf{TP} & \textbf{Loss} \\
 \midrule
InputXGradient ($\mu$) & .0707 $\pm$ .003 & .0902 $\pm$ .003 & .0605 $\pm$ .001 & .0746 $\pm$ .001 \\
InputXGradient ($L_2$) & .7308 $\pm$ .001 & .7317 $\pm$ .001 & .7374 $\pm$ .003 & .7372 $\pm$ .003 \\
Saliency ($\mu$) & .6570 $\pm$ .003 & .6578 $\pm$ .002 & .5234 $\pm$ .001 & .5226 $\pm$ .002 \\
Saliency ($L_2$) & .6574 $\pm$ .003 & .6583 $\pm$ .002 & .5202 $\pm$ .001 & .5199 $\pm$ .002 \\
GuidedBackprop ($\mu$) & .0034 $\pm$ .005 & .0010 $\pm$ .003 & .0028 $\pm$ .000 & .0082 $\pm$ .002 \\
GuidedBackprop ($L_2$) & .6574 $\pm$ .003 & .6583 $\pm$ .002 & .5202 $\pm$ .001 & .5199 $\pm$ .002 \\
IntegratedGrads ($\mu$) & .2082 $\pm$ .009 & .2887 $\pm$ .009 & .1968 $\pm$ .025 & .2163 $\pm$ .009 \\
IntegratedGrads ($L_2$) & .6274 $\pm$ .008 & .5676 $\pm$ .004 & .6321 $\pm$ .017 & .6040 $\pm$ .016 \\
Activation ($\mu$) & .7333 $\pm$ .001 & .7333 $\pm$ .001 & .4271 $\pm$ .001 & .4271 $\pm$ .001 \\
Activation ($L_2$) & .7234 $\pm$ .000 & .7234 $\pm$ .000 & .6857 $\pm$ .000 & .6857 $\pm$ .000 \\
LIME & .0668 $\pm$ .016 & .0945 $\pm$ .005 & .0168 $\pm$ .011 & .0420 $\pm$ .008 \\
Occlusion & .2568 $\pm$ .008 & .3422 $\pm$ .007 & .0678 $\pm$ .029 & .0929 $\pm$ .007 \\
Shapley & .3816 $\pm$ .047 & .4209 $\pm$ .040 & .0803 $\pm$ .011 & .1442 $\pm$ .034 \\
\bottomrule
 \end{tabular}
\caption{
Cross-lingual faithfulness results for the French split of XNLI dataset: Scores are measured for different attribution methods on the XNLI dataset and averaged across models trained with different seeds for each architecture. Attribution calculations are performed with respect to the top prediction (TP) score and the loss.
}
 \label{crosslingualfaithfulness-results-french}
 \end{table*}

 \begin{table*}[t]
\centering
\begin{tabular}{lrrrr}
\toprule
\multirow{3}[3]{*}{\textbf{Method}} & \multicolumn{4}{c}{\textbf{$\rho$}} \\
\cmidrule(lr){2-5}
 &  \multicolumn{2}{c}{\textbf{mBERT}} & \multicolumn{2}{c}{\textbf{XLM-R}} \\
 \cmidrule(lr){2-5}
 & \textbf{TP} & \textbf{Loss} & \textbf{TP} & \textbf{Loss} \\
 \midrule
InputXGradient ($\mu$) & .0024 $\pm$ .001 & .0051 $\pm$ .001 & .0076 $\pm$ .002 & .0174 $\pm$ .001 \\
InputXGradient ($L_2$) & .2924 $\pm$ .001 & .2953 $\pm$ .001 & .2749 $\pm$ .001 & .2761 $\pm$ .001 \\
Saliency ($\mu$) & .2166 $\pm$ .003 & .2196 $\pm$ .004 & .1712 $\pm$ .002 & .1702 $\pm$ .002 \\
Saliency ($L_2$) & .2168 $\pm$ .003 & .2198 $\pm$ .004 & .1705 $\pm$ .002 & .1692 $\pm$ .002 \\
GuidedBackprop ($\mu$) & .0007 $\pm$ .003 & .0007 $\pm$ .001 & .0017 $\pm$ .001 & .0025 $\pm$ .000 \\
GuidedBackprop ($L_2$) & .2168 $\pm$ .003 & .2198 $\pm$ .004 & .1705 $\pm$ .002 & .1692 $\pm$ .002 \\
IntegratedGrads ($\mu$) & .0286 $\pm$ .008 & .0758 $\pm$ .012 & .0963 $\pm$ .014 & .1173 $\pm$ .017 \\
IntegratedGrads ($L_2$) & .2597 $\pm$ .009 & .2433 $\pm$ .009 & .2059 $\pm$ .008 & .1970 $\pm$ .008 \\
Activation ($\mu$) & .2462 $\pm$ .000 & .2462 $\pm$ .000 & .1307 $\pm$ .000 & .1307 $\pm$ .000 \\
Activation ($L_2$) & .2127 $\pm$ .000 & .2127 $\pm$ .000 & .2007 $\pm$ .000 & .2007 $\pm$ .000 \\
LIME & .0281 $\pm$ .022 & .0173 $\pm$ .008 & .0041 $\pm$ .002 & .1552 $\pm$ .003 \\
Occlusion & .0451 $\pm$ .011 & .0591 $\pm$ .009 & .0128 $\pm$ .003 & .0305 $\pm$ .002 \\
Shapley & .3001 $\pm$ .063 & .2461 $\pm$ .051 & .0283 $\pm$ .013 & .0741 $\pm$ .015 \\
\bottomrule
 \end{tabular}
\caption{
Cross-lingual faithfulness results for the Thai split of XNLI dataset: Scores are measured for different attribution methods on the XNLI dataset and averaged across models trained with different seeds for each architecture. Attribution calculations are performed with respect to the top prediction (TP) class and the loss.
}
 \label{crosslingualfaithfulness-results-thai}
 \end{table*}
 
 \begin{table*}[t]
\centering
\begin{tabular}{lrrrr}
\toprule
\multirow{3}[3]{*}{\textbf{Method}} & \multicolumn{4}{c}{\textbf{$\rho$}} \\
\cmidrule(lr){2-5}
 &  \multicolumn{2}{c}{\textbf{mBERT}} & \multicolumn{2}{c}{\textbf{XLM-R}} \\
 \cmidrule(lr){2-5}
 & \textbf{TP} & \textbf{Loss} & \textbf{TP} & \textbf{Loss} \\
 \midrule
InputXGradient ($\mu$) & .0064 $\pm$ .001 & .0132 $\pm$ .002 & .0167 $\pm$ .002 & .0253 $\pm$ .003 \\
InputXGradient ($L_2$) & .4598 $\pm$ .001 & .4616 $\pm$ .001 & .5254 $\pm$ .002 & .5255 $\pm$ .002 \\
Saliency ($\mu$) & .4083 $\pm$ .002 & .4107 $\pm$ .002 & .4136 $\pm$ .002 & .4132 $\pm$ .002 \\
Saliency ($L_2$) & .4084 $\pm$ .002 & .4108 $\pm$ .002 & .4110 $\pm$ .002 & .4104 $\pm$ .002 \\
GuidedBackprop ($\mu$) & .0043 $\pm$ .002 & -.0015 $\pm$ .003 & .0002 $\pm$ .003 & .0015 $\pm$ .001 \\
GuidedBackprop ($L_2$) & .4084 $\pm$ .002 & .4108 $\pm$ .002 & .4110 $\pm$ .002 & .4104 $\pm$ .002 \\
IntegratedGrads ($\mu$) & .0713 $\pm$ .005 & .1239 $\pm$ .006 & .1137 $\pm$ .021 & .1223 $\pm$ .005 \\
IntegratedGrads ($L_2$) & .3806 $\pm$ .005 & .3298 $\pm$ .005 & .4883 $\pm$ .009 & .4634 $\pm$ .011 \\
Activation ($\mu$) & .4686 $\pm$ .001 & .4686 $\pm$ .001 & .3400 $\pm$ .001 & .3400 $\pm$ .001 \\
Activation ($L_2$) & .4987 $\pm$ .000 & .4987 $\pm$ .000 & .5449 $\pm$ .000 & .5449 $\pm$ .000 \\
LIME & .0257 $\pm$ .015 & .0752 $\pm$ .002 & .0128 $\pm$ .005 & .1854 $\pm$ .001 \\
Occlusion & .0537 $\pm$ .003 & .0988 $\pm$ .001 & .0416 $\pm$ .016 & .0628 $\pm$ .010 \\
Shapley & .2424 $\pm$ .044 & .2622 $\pm$ .044 & .0612 $\pm$ .015 & .1174 $\pm$ .014 \\
\bottomrule
 \end{tabular}
\caption{
Cross-lingual faithfulness results for the Swahili split of XNLI dataset: Scores are measured for different attribution methods on the XNLI dataset and averaged across models trained with different seeds for each architecture. Attribution calculations are performed with respect to the top prediction (TP) class and the loss.
}
 \label{crosslingualfaithfulness-results-swahili}
 \end{table*}
 
 \begin{table*}[t]
\centering
\begin{tabular}{lrrrr}
\toprule
\multirow{3}[3]{*}{\textbf{Method}} & \multicolumn{4}{c}{\textbf{$\rho$}} \\
\cmidrule(lr){2-5}
 &  \multicolumn{2}{c}{\textbf{mBERT}} & \multicolumn{2}{c}{\textbf{XLM-R}} \\
 \cmidrule(lr){2-5}
 & \textbf{TP} & \textbf{Loss} & \textbf{TP} & \textbf{Loss} \\
 \midrule
InputXGradient ($\mu$) & .0156 $\pm$ .003 & .0211 $\pm$ .005 & .0147 $\pm$ .002 & .0225 $\pm$ .004 \\
InputXGradient ($L_2$) & .5522 $\pm$ .002 & .5533 $\pm$ .002 & .5031 $\pm$ .003 & .5037 $\pm$ .003 \\
Saliency ($\mu$) & .4492 $\pm$ .004 & .4509 $\pm$ .004 & .2512 $\pm$ .004 & .2513 $\pm$ .004 \\
Saliency ($L_2$) & .4499 $\pm$ .004 & .4515 $\pm$ .004 & .2495 $\pm$ .004 & .2496 $\pm$ .004 \\
GuidedBackprop ($\mu$) & -.0003 $\pm$ .003 & .0006 $\pm$ .001 & -.0017 $\pm$ .003 & .0011 $\pm$ .006 \\
GuidedBackprop ($L_2$) & .4499 $\pm$ .004 & .4515 $\pm$ .004 & .2495 $\pm$ .004 & .2496 $\pm$ .004 \\
IntegratedGrads ($\mu$) & .0700 $\pm$ .003 & .1453 $\pm$ .003 & .0886 $\pm$ .009 & .1187 $\pm$ .008 \\
IntegratedGrads ($L_2$) & .4451 $\pm$ .012 & .3909 $\pm$ .008 & .4168 $\pm$ .012 & .3955 $\pm$ .014 \\
Activation ($\mu$) & .4700 $\pm$ .000 & .4700 $\pm$ .000 & .2407 $\pm$ .001 & .2407 $\pm$ .001 \\
Activation ($L_2$) & .5688 $\pm$ .000 & .5688 $\pm$ .000 & .4820 $\pm$ .000 & .4820 $\pm$ .000 \\
LIME & .0398 $\pm$ .013 & .0593 $\pm$ .003 & .0049 $\pm$ .003 & .1077 $\pm$ .006 \\
Occlusion & .0815 $\pm$ .005 & .1382 $\pm$ .006 & .0037 $\pm$ .012 & .0213 $\pm$ .013 \\
Shapley & .2557 $\pm$ .044 & .2915 $\pm$ .038 & .0229 $\pm$ .008 & .0490 $\pm$ .009 \\
\bottomrule
 \end{tabular}
\caption{
Cross-lingual faithfulness results for the Urdu split of XNLI dataset: Scores are measured for different attribution methods on the XNLI dataset and averaged across models trained with different seeds for each architecture. Attribution calculations are performed with respect to the top prediction (TP) class and the loss.
}
 \label{crosslingualfaithfulness-results-urdu}
 \end{table*}

 \section{Human Evaluation for e-XNLI}
\label{sec:human-evaluation-e-xnli}

A subset of our dataset is evaluated by NLP researchers---the authors and a colleague of one of the authors---from Turkey, Israel, and Russia.

The annotators followed the e-SNLI guidelines specified in Section 3 of \citet{Camburu2018eSNLINL} for evaluating whether automatically-extracted highlight-based explanations are correct. Note that incorrectly predicted examples are ignored during the evaluation.

 \section{Computational Resources}
\label{sec:computational-resources}

We mainly used Google Colab for the experiments and Titan RTX in some cases. All experiments for gradient-based attribution methods and Activation take a period of time ranging from 5 minutes to 1 hour, while perturbation-based approaches take several hours. Especially, experiments for Shapley Value Sampling take a few days since its implementation does not use batched operations.

\end{document}